\documentclass[acmsmall,screen]{acmart}
\usepackage{xcolor}

\AtBeginDocument{%
  \captionsetup{font=scriptsize,labelfont=bf}%
  \captionsetup[figure]{font=scriptsize,labelfont=bf}%
  \captionsetup[table]{font=scriptsize,labelfont=bf}%
}

\setcopyright{acmlicensed}
\copyrightyear{2026}
\acmYear{2026}
\acmDOI{XXXXXXX.XXXXXXX}
\acmJournal{CSUR}
\acmVolume{1}
\acmNumber{1}
\acmArticle{1}
\acmMonth{9}
\usepackage{float}
\usepackage{booktabs}
\usepackage{multirow}
\usepackage{array}
\usepackage{tabularx}
\usepackage{longtable}
\usepackage{amsfonts}
\usepackage{amsmath}
\usepackage[table]{xcolor}
\usepackage{siunitx}

\usepackage{tikz}
\usepackage{forest}
\usepackage{microtype}
\usepackage{wasysym}
\usepackage[most]{tcolorbox}
\usepackage{pdflscape}
\usepackage{enumitem}
\usepackage{multicol}
\setlist{nosep, leftmargin=*, topsep=1pt, itemsep=1pt}
\makeatletter
\renewcommand\fps@figure{tbp}
\renewcommand\fps@table{tbp}
\makeatother
\AtBeginDocument{%
}

\newtcolorbox{takeaway}[1][]{
  enhanced,
  breakable,
  colback=takeawaybg,
  colframe=takeawayframe,
  boxrule=0.3pt,
  arc=3pt,
  boxsep=0pt,
  left=2pt, right=2pt, top=1pt, bottom=1pt,
  before skip=2pt,
  after skip=2pt,
  fonttitle=\bfseries\tiny,
  fontupper=\scriptsize,
  coltitle=white,
  attach boxed title to top left={xshift=5pt, yshift=-\tcboxedtitleheight/4},
  boxed title style={
    colback=takeawaytab,
    colframe=takeawaytab,
    boxrule=0pt,
    arc=1.0pt,
    left=0.5pt, right=0.5pt, top=0pt, bottom=0pt,
  },
  title={Takeaway~\thetcbcounter},
  #1
}

\usetikzlibrary{arrows.meta, positioning, shapes, shapes.geometric, backgrounds, fit, calc, external}

\newcommand{\leadin}[1]{\textbf{\textit{#1}}}
\newcommand{\leadinB}[1]{{\textit{#1}}}

\definecolor{problemA}{RGB}{235,245,255}
\definecolor{problemB}{RGB}{235,250,240}
\definecolor{problemC}{RGB}{255,245,230}
\definecolor{problemD}{RGB}{245,240,255}
\definecolor{rootgray}{RGB}{220,218,210}
\definecolor{rootgraydark}{RGB}{90,88,82}
\definecolor{purplelight}{RGB}{206,203,246}
\definecolor{purpledark}{RGB}{60,52,137}
\definecolor{purplemid}{RGB}{143,137,221}
\definecolor{teallight}{RGB}{159,225,203}
\definecolor{tealdark}{RGB}{8,80,65}
\definecolor{tealmid}{RGB}{29,158,117}
\definecolor{bluemid}{RGB}{100,150,255}
\definecolor{amberlight}{RGB}{250,199,117}
\definecolor{amberdark}{RGB}{99,56,6}
\definecolor{ambermid}{RGB}{186,117,23}
\definecolor{corallight}{RGB}{245,196,179}
\definecolor{coraldark}{RGB}{113,43,19}
\definecolor{pinklight}{RGB}{244,192,209}
\definecolor{pinkdark}{RGB}{114,36,62}
\definecolor{paperbox}{RGB}{240,238,232}
\definecolor{paperborder}{RGB}{180,178,169}
\definecolor{bluelight}{HTML}{60A5FA}
\definecolor{bluedark}{HTML}{1D4ED8}
\definecolor{greenlight}{HTML}{34D399}
\definecolor{greendark}{HTML}{065F46}
\definecolor{graylight}{RGB}{241,239,232}
\definecolor{graydark}{RGB}{68,68,65}
\definecolor{grayborder}{RGB}{180,178,169}
\definecolor{takeawaybg}{HTML}{FDF3E7}
\definecolor{takeawaytab}{HTML}{B8874A}
\definecolor{takeawayframe}{HTML}{B8874A}

\DeclareRobustCommand{\fullcircle}{%
  \tikz[baseline=-0.85ex]{%
    \draw[fill=black, line width=0.3pt] (0,0) circle (0.85ex);%
  }%
}

\DeclareRobustCommand{\halfcircle}{%
  \tikz[baseline=-0.85ex]{%
    \draw[line width=0.3pt] (0,0) circle (0.85ex);%
    \begin{scope}
      \clip (0,0) circle (0.85ex);
      \fill (-0.85ex,-0.85ex) rectangle (0,0.85ex);
    \end{scope}%
  }%
}

\DeclareRobustCommand{\emptycircle}{%
  \tikz[baseline=-0.85ex]{%
    \draw[line width=0.3pt] (0,0) circle (0.85ex);%
  }%
}
\newcolumntype{P}[1]{>{\centering\arraybackslash}p{#1}}

\newcommand{\finding}[2]{%
  \par\medskip\noindent
  \fcolorbox{black!40}{gray!10}{%
    \parbox{\dimexpr\linewidth-2\fboxsep-2\fboxrule\relax}{%
      \small\textbf{Finding #1.} #2}}%
  \par\medskip}

\newcommand{\inlineeqnum}[1]{%
  \refstepcounter{equation}\label{#1}\hfill\mbox{(\theequation)}\linebreak}

\begin{document}

\title[Rollout Efficiency in RL for Reasoning LLMs]%
      {Rollout Efficiency in Reinforcement Learning for Reasoning Large
       Language Models: A Taxonomy and Future Directions}


\author{Niloofar Gholipour}
\email{niloofar.gholipour.1@ens.etsmtl.ca}
\author{Marcos Assuncao}
\email{marcos.dias-de-assuncao@etsmtl.ca}
\affiliation{%
  \institution{École de technologie supérieure, Univ. of Québec}
  \city{Montréal}
  \country{Canada}
}

\author{Gursimran Singh}
\authornote{Gursimran Singh and Timothy Yu contributed equally to this research.}
\email{gursimran.singh1@huawei.com}
\author{Timothy Yu}
\authornotemark[1]
\email{timothy.yu@huawei.com}
\affiliation{%
  \institution{Huawei Technologies}
  \city{Vancouver}
  \country{Canada}
}

\author{Rajkumar Buyya}
\email{rbuyya@unimelb.edu.au}
\affiliation{%
  \institution{The Univ. of Melbourne}
  \city{Melbourne}
  \country{Australia}
}

\author{Julien Gascon-Samson}
\email{julien.gascon-samson@etsmtl.ca}
\affiliation{%
  \institution{École de technologie supérieure, Univ. of Québec}
  \city{Montréal}
  \country{Canada}
}

\author{Zhenan Fan}
\email{zhenan.fan1@huawei.com}
\author{Yong Zhang}
\email{yong.zhang3@huawei.com}
\affiliation{%
  \institution{Huawei Technologies}
  \city{Vancouver}
  \country{Canada}
}

\author{xiaojie xu}
\email{xuxiaojie5@huawei.com}
\author{yaqiang yao}
\email{yaoyaqiang@huawei.com}
\author{xiaolong bai}
\email{baixiaolong1@huawei.com}
\affiliation{%
  \institution{Huawei Technologies}
  \country{China}
}

\authorsaddresses{Authors' addresses: N.~Gholipour, M.~Assuncao, and
J.~Gascon-Samson, \'Ecole de technologie sup\'erieure, Canada,
niloofar.gholipour.1@ens.etsmtl.ca,
\{marcos.dias-de-assuncao, julien.gascon-samson\}@etsmtl.ca;
G.~Singh, T.~Yu, Z.~Fan, and Y.~Zhang, Huawei Technologies, Canada,
\{gursimran.singh1, timothy.yu, zhenan.fan1, yong.zhang3\}@huawei.com;
R.~Buyya, The Univ.\ of Melbourne, Australia, rbuyya@unimelb.edu.au.}

\renewcommand{\shortauthors}{N. Gholipour et al.}
\begin{abstract}
Reasoning-oriented reinforcement learning enables large language models
to solve mathematical, coding, and other multi-step tasks, but shifts a
substantial portion of the training cost to rollout, where trajectories
are generated for policy updates. Efficient rollout mechanisms are
therefore essential to reduce this cost while maintaining the freshness,
consistency, and statistical validity of training data. This survey
provides a systematic taxonomy of recent research on rollout efficiency
for reasoning-oriented reinforcement learning, classifying existing
approaches from both mechanism and bottleneck perspectives. Based on this
taxonomy, we analyze how different technique families address distinct
sources of rollout inefficiency, examine opportunities and potential
conflicts for combining them, identify gaps in the evaluation and
reporting of efficiency gains, and discuss open challenges and future
research directions.
\end{abstract}



\maketitle

\section{Introduction}
\label{sec:introduction}

Difficult reasoning tasks require large language models (LLMs) to explore
possible solutions, decompose problems, correct mistakes, and adapt the amount
of computation they use. Pre-training provides broad knowledge, while
supervised fine-tuning (SFT) teaches these behaviors by imitating
demonstrations. Its effectiveness, however, depends on the availability,
coverage, and quality of those demonstrations. Reinforcement learning (RL)
offers a complementary approach: models generate alternative reasoning
trajectories and learn directly from task-level outcomes that indicate which
ones succeed. Although RL-based post-training also supports alignment and
safety, this survey focuses on reasoning, where repeatedly generating
trajectories during training creates a major computational cost. Many of the
techniques surveyed nevertheless apply to other RL settings that use iterative
trajectory generation. This shift is a central ingredient of recent reasoning-oriented LLMs. OpenAI's
o1 showed that large-scale RL teaches models to reason through extended
chains of thought and improve systematically with additional train-time and
test-time compute~\cite{jaech2024openai}; Kimi k1.5 showed that scaling RL
substantially improves reasoning across mathematics, coding, and other
challenging tasks~\cite{team2025kimi}; and DeepSeek-R1 showed that
outcome-based RL with verifiable rewards induces sophisticated behaviors such
as self-reflection and strategy adaptation while reducing reliance on
human-annotated reasoning trajectories~\cite{guo2025deepseek}.
Unlike SFT, which optimizes on a fixed set of demonstrations, RL generates
part of its training data online. For each batch of tasks, the current
\emph{policy}---the model being trained---samples one or more
\emph{trajectories}, which are scored by a reward signal and used to update
the policy; the updated policy then generates the next batch, forming a
repeated \emph{generate--evaluate--update} loop. This introduces a
computational bottleneck absent from pre-training and SFT, where most compute
goes to forward and backward passes over fixed data: substantial compute is
spent on \emph{rollout generation}, and because the policy changes with every
update, trajectories must be regenerated repeatedly rather than reused.
\begin{figure*}[h]
    \centering
    \captionsetup{labelfont={bf,sf,scriptsize},textfont={sf,scriptsize}}
    \includegraphics[width=\linewidth]{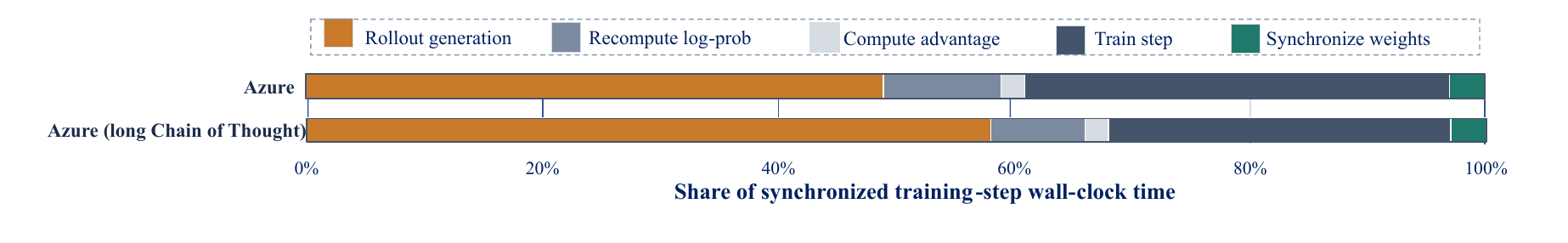}
    \caption{\textbf{Phase-time breakdown of a synchronous RL training step.}}
    \label{fig:rollout-dominates}
\end{figure*}
This generation cost can dominate the wall-clock time of an RL step. Under
\emph{synchronous} training, where the policy update waits for the required
trajectories to finish, prior measurements attribute roughly
70\%~\cite{gao2025rollpacker}, 85\%~\cite{hu2026taming}, and over 90\% in
long-output workloads~\cite{fu2026areal,zhou2025april} of step time to
rollout. Autoregressive decoding contributes directly, while variation in
trajectory length creates stragglers that further delay the synchronization
barrier~\cite{han2025asyncflow}. Our own synchronous measurements on
production inference traces show the same trend at smaller scale: over the
first 60 training steps, rollout is the largest single phase, at 49\% of mean
step time on the base workload and 58\% on its long-chain-of-thought variant
(Figure~\ref{fig:rollout-dominates}; setup in
Appendix~\ref{app:measurement}). The lower share relative to the studies
above reflects the shorter production-style responses in the base trace, and
its increase as outputs lengthen shows how strongly rollout cost depends on
the workload. More generally, the rollout fraction varies with model scale,
task, group size, trajectory-length distribution, reward-evaluation cost, and
execution design. Rollout is generated with the same inference engines used for deployment, but
it is not ordinary serving. A rollout produces the training data that updates
the model, so it sits on the critical path of every iteration, its slowest
trajectories rather than its average ones set the iteration
time~\cite{fu2026areal,han2025asyncflow}, and its outputs---including
token-level log-probabilities---must agree with the computation used in
training, a requirement we call \emph{rollout--training
consistency}~\cite{zhong2026diagnosing}. Rollout efficiency therefore means
reducing the cost of generating useful training trajectories while preserving
the conditions for valid policy updates.
Section~\ref{subsec:rollout-vs-inference} develops this contrast in detail.

\subsection{Challenges of Rollout Efficiency}
\label{subsec:challenges}

Because rollouts are the data that update the policy, changing how they are
generated can change the training signal. The main challenges are: ~(1)\leadinB{Policy lag and stale data.} Overlapping rollout generation with optimization increases throughput but
    lets trajectories from older policy versions enter later updates, creating
    a gap between the policy that generated the data and the policy being
    optimized that requires staleness control or off-policy
    correction~\cite{fu2026areal,han2025asyncflow}.~(2)\leadinB{Length dispersion and synchronization bubbles.}
    Reasoning trajectories vary substantially in length. Under synchronous
    execution, a few long trajectories determine the rollout makespan while
    workers that finish early sit idle at the barrier
    \cite{gao2025rollpacker,fu2026areal}. ~(3)\leadinB{Evolving length distributions.}
    Trajectory lengths change as the policy evolves, so fixed scheduling,
    batching, and resource-allocation decisions become inefficient as
    training progresses~\cite{he2025history}.~(4)\leadinB{Rollout--training mismatch.}
    The rollout and training engines may assign different log-probabilities to
    the same tokens despite identical weights, silently distorting the
    quantities the optimizer relies on~\cite{zhong2026diagnosing}.~(5)\leadinB{Sample efficiency and trajectory selection.}
    Trajectories contribute unequally to policy improvement, so giving every
    prompt the same rollout budget or retaining every trajectory spends
    generation compute on samples with little learning
    signal~\cite{zhang2026aero}.

\subsection{Motivation and Scope}
\label{subsec:motivation}

Rollout is a major bottleneck in reasoning RL, yet no survey treats it as a
single cost-to-target-quality problem spanning both execution and learning
efficiency. Existing surveys cover reasoning RL, inference efficiency,
rollout strategies, or distributed training from different perspectives
(Section~\ref{sec:related-surveys}), each addressing one side of that cost
or the other, but not jointly. The work that addresses this cost is split
across the systems and algorithms communities, which optimize the same
rollout budget from different directions and are often deployed together.

\leadin{Rollout efficiency.}
We frame \emph{rollout efficiency} as the cost required to reach a target
level of policy quality: a method improves it when it lowers that cost. Two
complementary levers do so. The \emph{system lever} lowers the cost of
executing a given rollout workload through scheduling, asynchronous
execution, and decoding acceleration. The \emph{algorithmic lever} lowers the
rollout work required for learning through prompt filtering, adaptive sample
allocation, and truncation. Faster decoding makes a trajectory cheaper but
does not make it useful for learning; filtering reduces the work required
but does not make the remaining trajectories cheaper. A method can therefore
affect either lever or both, but the literature rarely measures both: of
the 80 methods we compare, only 12 report a gain on both levers
(Section~\ref{sec:metrics}). This makes overall efficiency hard to compare
across studies and motivates the evaluation analysis later in the survey.
The two levers form the primary axis of our taxonomy, and the operational
view in Section~\ref{sec:background} locates where each intervention acts in
the training loop. Our scope is rollout efficiency in trajectory collection
during training; test-time inference efficiency and purely statistical data
efficiency are adjacent topics covered by prior surveys and discussed only
where they intersect rollout cost.

\subsection{Research Questions and Contributions}
\label{subsec:contributions}

This survey is structured around four research questions:

\begin{itemize}
\item \leadin{RQ1:} Where does rollout cost arise in the reasoning-RL training
loop, and which workload properties make it persistent?
(Section~\ref{sec:background})
\item \leadin{RQ2:} By what mechanisms do existing methods reduce this cost, and
how do these mechanisms relate to one another?
(Sections~\ref{sec:taxonomy}--\ref{sec:algorithm})
\item \leadin{RQ3:} Which techniques can be combined safely, and which
combinations compound staleness, numerical mismatch, or statistical bias?
(Section~\ref{sec:composability})
\item \leadin{RQ4:} How should rollout-efficiency gains be measured so that
results from different studies become comparable?
(Section~\ref{sec:metrics})
\end{itemize}
The survey makes four contributions: (1)~We organize rollout-efficiency methods into a dual taxonomy by
\emph{how} they improve efficiency and by \emph{which rollout bottleneck}
they address, and analyze each family's design space, validity conditions,
and applicable settings.
(2)~We provide a composability analysis of which technique families can be
combined, which combinations are promising, and which interfere with each
other.
(3)~We give a measured account of reporting practice, showing that system
efficiency and learning efficiency are rarely measured together, and unify
the efficiency measures used across the literature into a common
cost-to-quality view with a minimum reporting set and diagnostic protocols.
(4)~We identify open problems and future directions drawn from the gaps the
taxonomy, composability analysis, and evaluation study expose.

\leadin{Article Organization.}Section~\ref{sec:related-surveys} positions our scope relative to prior surveys;
Section~\ref{sec:background} establishes the rollout-efficiency framework and
problem setting; Sections~\ref{sec:taxonomy}--\ref{sec:algorithm} develop the
dual taxonomy; Section~\ref{sec:composability} examines cross-family
interactions; Section~\ref{sec:metrics} develops the common evaluation lens;
and Section~\ref{sec:future_direction} discusses open problems and future
directions. Section~\ref{sec:summary} concludes. Appendix~\ref{subsec:methodology} documents the survey methodology: the search, screening, and classification protocol that produced the included works.

\section{Related Surveys}
\label{sec:related-surveys}

Prior surveys cover efficient training, reasoning, reinforcement learning,
and inference, but none takes the cost of generating training trajectories
as its unit of analysis; Table~\ref{tab:related-survey-comparison} compares
their scope with ours. Bai et al.~\cite{bai2401beyond} and Duan et
al.~\cite{duan2024efficient} review resource-efficient and distributed
training, where the unit is model computation for a fixed workload rather
than a policy-evolving loop that repeatedly creates its own data. Tie et
al.~\cite{tie2025survey} and Kumar et al.~\cite{kumar2025llm} survey
post-training and reasoning methods without isolating
trajectory-generation cost. Qu et al.~\cite{qu2025survey}, Liu et
al.~\cite{liu2025efficient}, Ke et al.~\cite{ke2025survey}, and Chen et
al.~\cite{chen2026towards} cover long chain-of-thought, test-time scaling,
and serving-time acceleration; several mechanisms overlap with ours, but
they treat efficiency as a property of a deployed model, whereas rollout
generation sits inside a loop where trajectories become training data and
the policy itself changes (Section~\ref{subsec:rollout-vs-inference}). Liu
et al.~\cite{liu2025reinforcement}, Zhang et al.~\cite{zhang2025survey},
Zhang et al.~\cite{zhang2509landscape}, and Wang et
al.~\cite{wang2025comprehensive} organize RL and agentic systems around
methodology, reward design, and architecture, and Li et
al.~\cite{li2026agentic} around environment design; none treats rollout cost
as a first-class dimension. Yu et al.~\cite{yu2026survey} and Zhang et
al.~\cite{zhang2026reasoning} study data-efficient RL and credit
assignment---how much learning a set of trajectories yields---rather than
what those trajectories cost to produce.
\begin{table}[H]
\centering
\captionsetup{labelfont={bf,sf,scriptsize},textfont={sf,scriptsize}}
\caption{\textbf{Comparison of representative surveys and this work.}
Scope states each survey's primary subject; parentheses mark a secondary
subject treated in passing. \fullcircle\ full coverage;
\halfcircle\ partial coverage; \emptycircle\ limited or no coverage.}
\label{tab:related-survey-comparison}
\scriptsize
\setlength{\tabcolsep}{3pt}
\renewcommand{\arraystretch}{0.85}
\begin{tabular}{l c l ccc ccc}
\toprule
\multirow{2}{*}{\textbf{Survey}} &
\multirow{2}{*}{\textbf{Year}} &
\multirow{2}{*}{\textbf{Scope}} &
\multicolumn{3}{c}{\textbf{Rollout-efficiency coverage}} &
\multicolumn{3}{c}{\textbf{Empirical analysis}} \\
\cmidrule(lr){4-6}
\cmidrule(lr){7-9}
& & &
\shortstack{System\\lever} &
\shortstack{Algorithmic\\lever} &
\shortstack{Bottleneck\\mapping} &
\shortstack{Reported\\gains} &
\shortstack{Compos-\\ability} &
\shortstack{Metric\\standardization} \\
\midrule
Bai et al.~\cite{bai2401beyond}         & 2024 & Training (+inference) &
  \halfcircle & \emptycircle & \emptycircle &
  \halfcircle & \emptycircle & \emptycircle \\
Duan et al.~\cite{duan2024efficient}    & 2024 & Training &
  \fullcircle & \emptycircle & \emptycircle &
  \halfcircle & \emptycircle & \emptycircle \\
Tie et al.~\cite{tie2025survey}         & 2025 & Training &
  \emptycircle & \emptycircle & \emptycircle &
  \emptycircle & \emptycircle & \emptycircle \\
Kumar et al.~\cite{kumar2025llm}        & 2025 & Training &
  \emptycircle & \emptycircle & \emptycircle &
  \emptycircle & \emptycircle & \emptycircle \\
Qu et al.~\cite{qu2025survey}           & 2025 & Inference &
  \emptycircle & \halfcircle & \emptycircle &
  \halfcircle & \emptycircle & \emptycircle \\
Liu et al.~\cite{liu2025efficient}      & 2025 & Inference &
  \halfcircle & \halfcircle & \emptycircle &
  \halfcircle & \emptycircle & \emptycircle \\
Ke et al.~\cite{ke2025survey}           & 2025 & Inference &
  \emptycircle & \emptycircle & \emptycircle &
  \emptycircle & \emptycircle & \emptycircle \\
Liu et al.~\cite{liu2025reinforcement}  & 2025 & Training (+inference) &
  \emptycircle & \emptycircle & \emptycircle &
  \emptycircle & \emptycircle & \emptycircle \\
Zhang et al.~\cite{zhang2025survey}     & 2025 & Training &
  \emptycircle & \emptycircle & \emptycircle &
  \halfcircle & \emptycircle & \emptycircle \\
Zhang et al.~\cite{zhang2509landscape}  & 2025 & Training &
  \emptycircle & \emptycircle & \emptycircle &
  \emptycircle & \emptycircle & \emptycircle \\
Wang et al.~\cite{wang2025comprehensive}& 2025 & Training (+inference) &
  \emptycircle & \emptycircle & \emptycircle &
  \emptycircle & \emptycircle & \emptycircle \\
Chen et al.~\cite{chen2026towards}      & 2026 & Inference &
  \emptycircle & \emptycircle & \emptycircle &
  \emptycircle & \emptycircle & \emptycircle \\
Li et al.~\cite{li2026agentic}          & 2026 & Training &
  \emptycircle & \emptycircle & \emptycircle &
  \emptycircle & \emptycircle & \emptycircle \\
Yu et al.~\cite{yu2026survey}           & 2026 & Training &
  \emptycircle & \fullcircle & \halfcircle &
  \halfcircle & \emptycircle & \emptycircle \\
Zhang et al.~\cite{zhang2026reasoning}  & 2026 & Training &
  \emptycircle & \halfcircle & \halfcircle &
  \halfcircle & \emptycircle & \emptycircle \\
Surana et al.~\cite{surana2026generate} & 2026 & Training &
  \emptycircle & \fullcircle & \halfcircle &
  \emptycircle & \emptycircle & \emptycircle \\
\midrule
\textbf{This survey}                    & 2026 & \textbf{Rollout} &
  \fullcircle & \fullcircle & \fullcircle &
  \fullcircle & \fullcircle & \fullcircle \\
\bottomrule
\end{tabular}
\end{table}
Closest to our scope, Surana et al.~\cite{surana2026generate} organize
rollout strategies through a Generate--Filter--Control--Replay pipeline, a
workflow-oriented view; ours is efficiency-oriented and cuts across workflow
stages, which is what lets system-level techniques (asynchronous execution,
scheduling, speculative decoding) be compared directly with algorithmic
ones (prompt filtering, adaptive allocation, trajectory selection). The
remaining columns of Table~\ref{tab:related-survey-comparison} mark the
other differences: no prior survey maps methods to the source of waste they
address, examines cross-family composability, or asks when reported gains
can be compared. We therefore do not claim that rollout strategies are
unsurveyed; we organize the literature around a different question---what
useful training progress costs to obtain from rollout generation, and how
that cost can be reduced and compared across methods.

\subsection{Survey Methodology}
\label{subsec:survey-methodology}

We build the survey corpus and taxonomy in four stages---\emph{search},
\emph{screen}, \emph{classify}, and \emph{group}---summarized in
Figure~\ref{fig:methodology}; the search query, full inclusion/exclusion
criteria, and edge-case rulings are given in
Appendix~\ref{subsec:methodology}. Searching ACM Digital Library, IEEE
Xplore, arXiv, and OpenReview for work published between January~2024 and
September~2026, supplemented by venue scans, snowballing, and framework
technical reports, we screened 150 candidate records after de-duplication
against three inclusion and three exclusion criteria (training-time, LLM
reasoning-RL, changes rollout cost); 86 are included, of which 80 receive
method-level rows in Table~\ref{tab:comparison}.

\begin{figure}[H]
\centering
\scriptsize
\captionsetup{labelfont={bf,sf,scriptsize},textfont={sf,scriptsize}}
\includegraphics[width=0.85\linewidth]{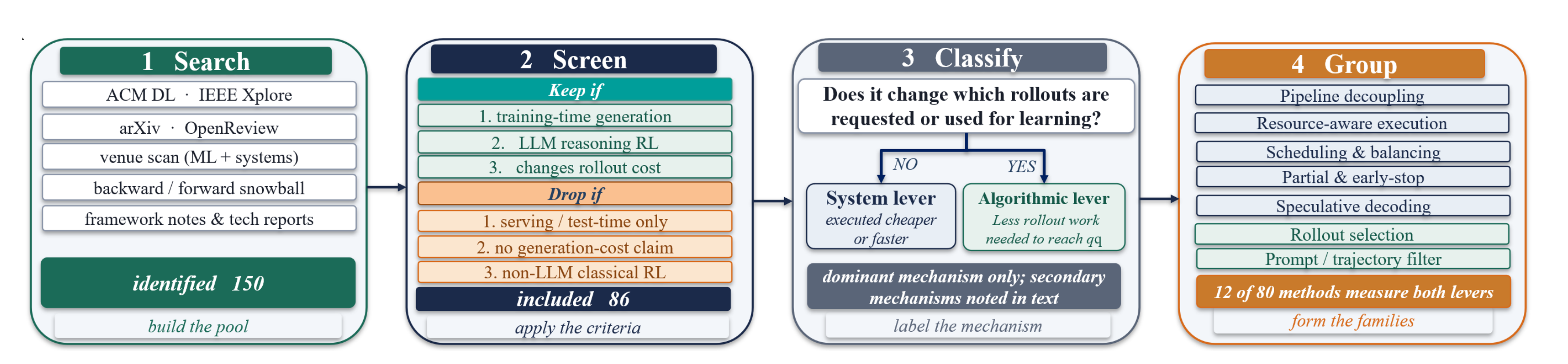}
\caption{\textbf{Survey methodology.} The taxonomy is constructed through four stages:
search, screen, classify, and group.}
\label{fig:methodology}
\end{figure}

Each work is assigned one primary mechanism family and one primary
bottleneck. The dominant lever is determined by where the intervention acts: methods
that make an already-requested rollout workload cheaper or faster take the
\emph{system lever}; methods that change which rollouts are requested or
which trajectories are used for learning take the \emph{algorithmic lever}.
The resulting seven mechanism families are cross-referenced with the four
bottlenecks of Section~\ref{sec:taxonomy}, and for each work we record the
reported efficiency measure, evaluation setting, and quality outcome used in
the analysis of Sections~\ref{sec:metrics} and~\ref{sec:composability}. The
taxonomy reflects each work's dominant mechanism as classified here; reported
speedups and compatibility claims remain tied to the conditions of the
original studies.

\section{Background and Rollout Efficiency Framework}
\label{sec:background}
\subsection{Conceptual Workflow of Reasoning RL}
\label{subsec:conceptual-workflow}
\begin{figure*}[h]
    \centering
    \scriptsize
    \captionsetup{labelfont={bf,sf,scriptsize},textfont={sf,scriptsize}}
    \includegraphics[width=\textwidth]{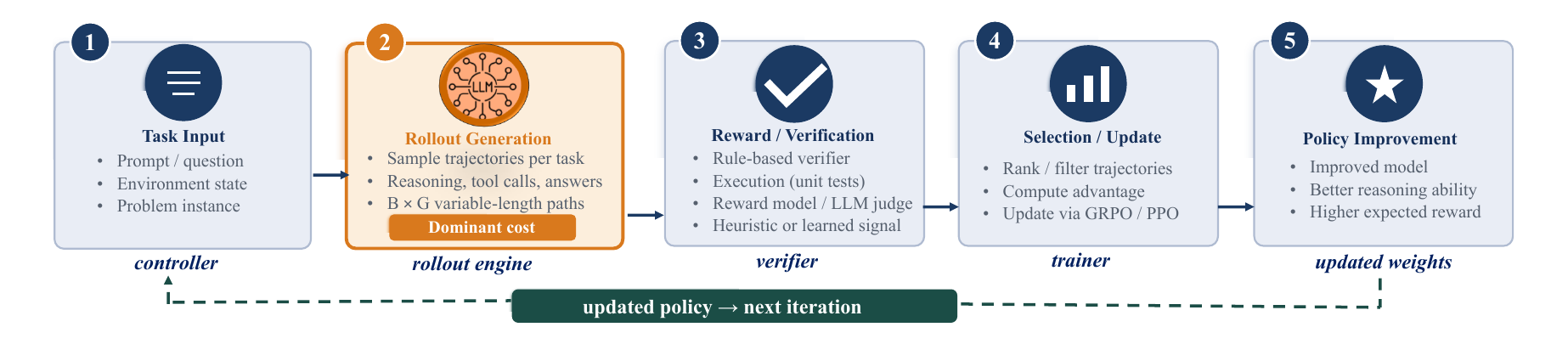}
    \caption{\textbf{Conceptual workflow of reasoning RL.}}
    \label{fig:conceptualframework}
\end{figure*}

The object of training is a policy $\pi_\theta$ (an LLM): a probability
distribution over the next action/token conditioned on the current context. In
reasoning RL, the sequence generated by the policy is a
\emph{trajectory}---reasoning steps, optional tool calls, and a final answer.
The objective is to improve the expected reward of $\pi_\theta$, which on
reasoning tasks corresponds to higher solution accuracy, more reliable tool
use, or more consistent multi-step deduction
\cite{guo2025deepseek,team2025kimi,jaech2024openai}. Unlike supervised
fine-tuning, where the training examples are supplied in advance, RL generates
part of its training data online: the policy produces trajectories, those
trajectories are evaluated, and the resulting learning signal updates the
policy. The updated policy then generates the next batch, forming a repeated
\emph{generate--evaluate--update} loop. Figure~\ref{fig:conceptualframework}
shows this closed-loop structure. Implementations may combine, split, reorder,
or overlap these stages, but the high-level dependency remains the same.

\leadinB{Task input.}
Each iteration begins with a batch of $B$ tasks: prompts, problem instances,
or environment states. A controller samples the batch and dispatches it to the
rollout engines; $B$ therefore contributes directly to the generation
workload.

\leadinB{Rollout generation.}
The current policy samples $G$ completions for each task; $G$ is the group
size. This generation stage is the \emph{rollout}: it produces the
trajectories that become the training data for the current RL update. Each
completion for task $x_i$ is a trajectory $y_i^{(g)}$ of length
$|y_i^{(g)}|$ tokens, so one iteration produces $B \times G$ trajectories
containing $\sum_{i,g} |y_i^{(g)}|$ generated tokens. We refer to this
collection as the \emph{rollout workload}.
In standard reasoning tasks, a trajectory is a sequence of reasoning tokens
followed by a final answer. In tool-using or agentic RL the same loop
applies, but a trajectory may interleave model actions with environment
observations, $\tau = (x, a_1, o_1, a_2, o_2, \dots, a_H, o_H)$, where $H$
is the number of interaction turns, $a_h$ may be reasoning text, a tool
call, or an environment action, and $o_h$ is the resulting observation.
Rollout cost then includes the generated tokens $\sum_h |a_h|$, repeated
prefills as the context grows, and waiting time from tools or environments.
These interactions make trajectory cost more heterogeneous and can amplify
the long-tail behavior that later becomes important for rollout efficiency.

\leadinB{Reward and verification.}
A reward function $r:\mathcal{X}\times\mathcal{Y}\rightarrow\mathbb{R}$
assigns each trajectory a score. Common sources in reasoning RL are
verifiable rewards (RLVR) from a deterministic checker or executable
environment \cite{shao2024deepseekmath,guo2025deepseek}; learned or
model-based rewards when no programmatic checker exists
\cite{yuan2025selfrewardinglanguagemodels,
kwon2023rewarddesignlanguagemodels,li2025generalistrewardmodelsinside};
and human preference rewards (RLHF) \cite{ouyang2022training}. These
sources can be combined, and some settings score intermediate steps or
provide rewards during interaction rather than only at termination
\cite{mu2024rulebasedrewardslanguage,peng2025agenticrewardmodelingintegrating}.
The choice affects the source, granularity, and cost of the learning
signal, but not the central loop: trajectories are generated, evaluated,
and used to improve the policy.
Reward sparsity is especially relevant. A task may provide little useful
feedback until a trajectory reaches a final answer, so multiple samples
may be needed before rewards distinguish better behavior from worse. The
number of samples per task is therefore both a statistical design choice
and a major determinant of rollout cost.

\leadinB{Selection and update.}
Rewards are converted into advantages that determine how generated
trajectories contribute to the policy update. Two instantiations recur
later. Proximal policy optimization (PPO) estimates advantages with a
learned critic \cite{schulman2017proximal}; group relative policy
optimization (GRPO) forms them relative to the other samples generated for
the same prompt,
$\hat{A}_i^{(g)} = \bigl(r(x_i,y_i^{(g)})-\mu_{r,i}\bigr)/(\sigma_{r,i}+\epsilon)$,
where $\mu_{r,i}$ and $\sigma_{r,i}$ are the reward mean and standard
deviation within prompt $i$'s group and $\epsilon$ is a small constant
\cite{shao2024deepseekmath}. If every completion in a group receives the
same reward, $\sigma_{r,i}=0$ and all of the group's advantages vanish even
though the group has consumed its full generation budget. We call such a
group \emph{degenerate}; this observation motivates a substantial part of
the algorithmic-lever literature (Section~\ref{sec:algorithm}).
The trainer then updates $\pi_\theta$. PPO-style objectives compare the
policy being optimized with the policy that generated the rollout through
the per-token importance ratio
$\rho_{i,t}^{(g)}(\theta) =
\pi_\theta(y_{i,t}^{(g)} \mid x_i, y_{i,<t}^{(g)}) \,/\,
\pi_{\theta_{\mathrm{old}}}(y_{i,t}^{(g)} \mid x_i, y_{i,<t}^{(g)})$.
When the two policies coincide, $\rho=1$ and the update is on-policy.
When they differ, the batch exhibits \emph{policy lag}: a batch used at
step $t$ but generated by $\pi_{\theta_{t-k}}$ has lag $k$. Allowing
$k>0$ enables generation, verification, and optimization to overlap, but
introduces an off-policy component---the central trade-off underlying
asynchronous rollout systems
\cite{nair2015massively,mnih2016asynchronous,espeholt2018impala}.
Separately, even at lag zero, the rollout and training engines may compute
different token probabilities for the same trajectory because of
differences in numerical precision, kernels, or
implementation---a \emph{rollout--training mismatch} that can alter the
quantities used by the policy update \cite{zhong2026diagnosing}. The
update produces a new parameter vector $\theta'$, which is propagated
back to the rollout engines and used to generate the next batch.

\leadin{Problem formulation.}
As defined in Section~\ref{subsec:motivation}, \emph{rollout efficiency} is
the cost required to reach a target level of policy quality. Of the five
stages of Figure~\ref{fig:conceptualframework}, only task input carries
negligible direct execution cost, leaving four timed phases in one
synchronous iteration: (1)~rollout, at cost $T_{\mathrm{roll}}$, dispatches
the tasks and generates the $B \times G$ trajectories; (2)~reward
evaluation, at cost $T_{\mathrm{rew}}$, verifies or scores them;
(3)~selection and policy update, at cost $T_{\mathrm{upd}}$, computes
advantages and applies the optimizer step; and (4)~weight synchronization,
at cost $T_{\mathrm{sync}}$, propagates the updated policy to the rollout
engines. Under synchronous execution these phases serialize over the
iterations needed to reach the target,
$T_{\mathrm{total}} = \sum_{t=1}^{N(\mathcal{S})}
\bigl(T_{\mathrm{roll}}^{(t)} + T_{\mathrm{rew}}^{(t)} +
T_{\mathrm{upd}}^{(t)} + T_{\mathrm{sync}}^{(t)}\bigr)$,
where the iteration count $N(\mathcal{S})$ is itself a function of the
rollout strategy $\mathcal{S}$: changing which trajectories are generated,
how they are generated, or how they are used can change both the cost of an
iteration and the number of iterations required. Asynchronous systems
overlap phases and shorten the critical path but introduce policy lag, and
reward evaluation and weight synchronization are not universally
negligible---verifier execution, sandboxed tests, judge-model inference,
and weight transfer can all occupy the critical path.
The rollout-efficiency problem is to minimize the cost of a strategy
subject to a target quality:  $\min_{\mathcal{S}} \; \mathcal{C}_{\mathcal{S}}(Q^*)$,\inlineeqnum{eq:rollout-efficiency-problem}
where $\mathcal{S}$ specifies generation, sample allocation, scheduling,
phase overlap, synchronization, and hardware placement; $Q$ is a
task-level quality measure such as held-out accuracy; $Q^*$ is the level at
which strategies are compared; and $\mathcal{C}_{\mathcal{S}}(q)$ is the
cost strategy $\mathcal{S}$ spends to first reach quality $q$, written
$\mathcal{C}(q)$ when the strategy is clear. A cheaper strategy that never
reaches $Q^*$ is not more efficient. $\mathcal{C}$ may be wall-clock time,
accelerator-hours, generated tokens, or monetary cost; these are not
interchangeable, since a method may reduce tokens while increasing
wall-clock time. Two readings of ``rollout efficiency'' must also be kept
apart. \emph{Rollout-path cost} is the cost of trajectory collection
itself---the $T_{\mathrm{roll}}$ and $T_{\mathrm{rew}}$ terms of one
iteration---and is what the system lever most directly targets.
\emph{End-to-end cost-to-quality} is $\mathcal{C}(q)$, and is what
\emph{rollout efficiency} means throughout. This definition is what places
post-generation methods such as rollout selection in scope: they leave the
rollout-path cost of the current batch unchanged yet can still lower
$\mathcal{C}(q)$. Section~\ref{sec:metrics} shows that conflating the two
readings is a principal source of incomparability in the literature and
adopts accelerator-hours to $Q^*$ as the common measure.

\subsection{Rollout versus Deployment Inference}
\label{subsec:rollout-vs-inference}

Rollout generation and deployment inference often use the same
generation engines, such as vLLM~\cite{kwon2023efficient} or
SGLang~\cite{zheng2023sglang}, but they optimize different workloads.
Deployment serves externally arriving requests; RL rollout generates
training data under an iterative learning algorithm. Five properties
follow. (1)~\emph{Controlled batch arrival:} the trainer holds a batch
of $B$ tasks at the start of each rollout phase and can reorder,
partition, or allocate them, rather than reacting to a user-driven
request stream. (2)~\emph{Batch makespan rather than request latency:}
in synchronous RL the relevant objective is the time until sufficient
data exist for the next update, so a short trajectory helps little if
the batch is blocked by a few long ones; asynchronous RL relaxes that
barrier but introduces policy lag
\cite{fu2026areal}. (3)~\emph{Repeated and partially predictable
workloads:} tasks recur across iterations, so previous lengths,
rewards, or difficulty can guide later batching, allocation, or early
termination. (4)~\emph{An evolving model state:} the policy changes
between updates, so cached activations remain valid only for the
weights that produced them. (5)~\emph{Training-specific correctness:}
generated tokens, rewards, and log-probabilities become optimizer
inputs, so policy freshness and rollout--training consistency are part
of the efficiency problem rather than implementation details
\cite{zhong2026diagnosing,sheng2025hybridflow}.
Table~\ref{tab:rollout-vs-inference}
(Appendix~\ref{app:rollout-vs-inference}) expands each property. These differences make rollout a distinct optimization problem, not a
training-time instance of serving. They also explain why rollout can
exploit task-aware allocation and repeated trajectory information that
ordinary serving does not have, and why serving optimizations do not
transfer unchanged.

\section{Taxonomy Design}
\label{sec:taxonomy}

Section~\ref{subsec:motivation} introduced two complementary levers on
rollout efficiency: \emph{executing a requested rollout workload more
efficiently} (the system lever) or \emph{reducing the rollout work required
for a given amount of learning progress} (the algorithmic lever). This
section turns them into a taxonomy with two views of the same literature.
The \leadinB{Mechanism Taxonomy} (Section~\ref{subsec:tax-approach}) asks
\emph{what intervention a method performs}, refining the two levers into
seven technique families; the \leadinB{Bottleneck Taxonomy}
(Section~\ref{subsec:tax-problem}) asks \emph{where inefficiency arises},
organizing the same methods by the operational source of waste. Because
many systems combine several mechanisms, membership follows explicit rules
rather than paper titles or implementation complexity. (i)~In the Mechanism
Taxonomy, a method is assigned to the family responsible for its reported
efficiency gain. (ii)~In the Bottleneck Taxonomy, it is assigned to the
bottleneck that mechanism most directly targets. (iii)~Each method thus
receives one primary family and one primary bottleneck; secondary
mechanisms are discussed but create no additional memberships. (iv)~The
\emph{dominant} lever is determined by the control decision the
intervention changes: \emph{system} methods take the requested rollouts as
given and change how they are executed, whereas \emph{algorithmic} methods
change which rollouts are requested, how much budget they receive, or which
generated rollouts are used for learning. The distinction rests on the
decision being changed, not on the number of tokens ultimately generated; a
method that touches both dimensions is placed by the one that dominates its
reported effect.

\subsection{Mechanism Taxonomy --- Approaches to Rollout Efficiency}
\label{subsec:tax-approach}

\begin{figure}[t]
    \centering
    \captionsetup{labelfont={bf,sf,scriptsize},textfont={sf,scriptsize}}
    \makebox[\linewidth][c]{%
        \includegraphics[width=1.2\linewidth]{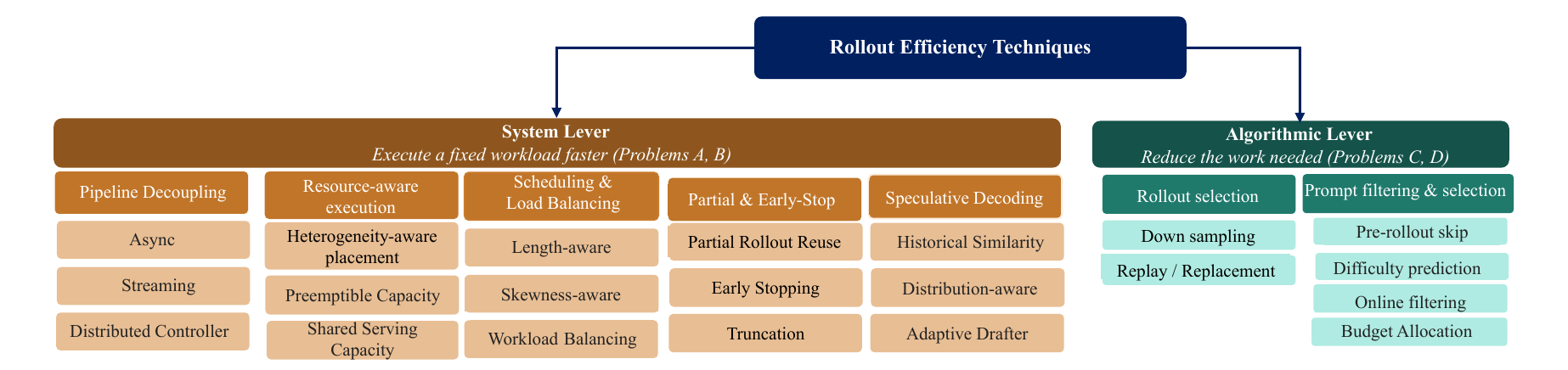}%
    }
    \caption{\textbf{Mechanism taxonomy of rollout-efficiency methods.}}
    \label{tax:mechanism}
\end{figure}

The Mechanism Taxonomy (Figure~\ref{tax:mechanism}) organizes methods by how
they reduce rollout cost, refining the two levers into seven technique
families and 21 mechanism variants. Table~\ref{tab:comparison-main} lists
two representative methods per family, chosen to illustrate the range of
system and algorithmic gains within each; the full comparison of all 80
surveyed methods, with scale, task, hardware, and cost/requirement details
for each, appears as Table~\ref{tab:comparison} in
Appendix~\ref{app:comparison}.

\begin{table}[!htb]
\centering
\scriptsize
\captionsetup{labelfont={bf,sf,scriptsize},textfont={sf,scriptsize}}
\renewcommand{\arraystretch}{0.85}
\setlength{\tabcolsep}{1.5pt}
\caption{\textbf{Representative methods per family} (2 of each; full
80-method comparison in Table~\ref{tab:comparison}, Appendix~\ref{app:comparison}).
Q: $=$ preserved, $+$ improved, $\approx$ comparable, \checkmark\ exact/lossless; --- not reported.}
\label{tab:comparison-main}
\begin{tabular}{|
>{\raggedright\arraybackslash}m{1.5cm}|
>{\raggedright\arraybackslash}m{2.1cm}|
>{\raggedright\arraybackslash}m{1.9cm}|
>{\centering\arraybackslash}m{1.0cm}|
>{\raggedright\arraybackslash}m{1.4cm}|
>{\raggedright\arraybackslash}m{1.2cm}|
>{\centering\arraybackslash}m{0.5cm}|
>{\raggedright\arraybackslash}m{2.5cm}|
}
\hline
\rowcolor{gray!15}
\textbf{Method} & \textbf{System gain} & \textbf{Algorithmic gain} &
\textbf{Scale} & \textbf{Task} & \textbf{Hardware} & \textbf{Q} &
\textbf{Cost / requirement} \\
\hline

\rowcolor{problemA}
\multicolumn{8}{|c|}{\textbf{A.1 Pipeline Decoupling}
\textnormal{\scriptsize(Problem~A: idle resources from stage coupling --- 2 of 11 shown)}}\\
\hline
AReaL~\cite{fu2026areal} & tput $2.77\times$ & --- & 1.5B--32B & math, code & --- & $=$/$+$ & stale trajectories within lag bound \\ \hline
LlamaRL~\cite{wu2025llamarl} & step $10.7\times$ (405B) & --- & 8B--405B & math & 1{,}024 H100 & $=$ & 1-step policy lag; AIPO correction \\ \hline

\rowcolor{problemA}
\multicolumn{8}{|c|}{\textbf{A.2 Resource-Aware Execution}
\textnormal{\scriptsize(Problem~A: idle, mismatched, or over-provisioned hardware --- 2 of 16 shown)}}\\
\hline
RLBoost~\cite{wu2025rlboost} & tput $1.51$--$1.97\times$ & --- & --- & --- & preempt.\ GPUs & --- & needs preemptible capacity; interruption recovery \\ \hline
AReaL-DTE~\cite{peng2026arealdte} & weight sync $6.8$--$7.6\times$ ($19.9\times$ cross-cluster) & --- & 8B; 30B-A3B & math, code, logic & 8--16 H200 & \checkmark & $<2\%$ weights change/step; AdamW inversion; periodic full anchors \\ \hline

\rowcolor{problemB}
\multicolumn{8}{|c|}{\textbf{B.1 Scheduling \& Load Balancing}
\textnormal{\scriptsize(Problem~B: long-tail rollout latency --- 2 of 7 shown)}}\\
\hline
RollPacker~\cite{gao2025rollpacker} & E2E $2.03$--$2.56\times$ & --- & 7B--32B & --- & $\leq$128 H800 & $=$ & length predictor accuracy; also $2.24\times$ vs.\ RLHFuse \\ \hline
SortedRL~\cite{zhang2026sortedrl} & bubble $-50\%$+ & $+3.9$--$18.4\%$ @ fixed data & 8B--32B & --- & --- & $+$ & length-sorted batches change the training distribution \\ \hline

\rowcolor{problemB}
\multicolumn{8}{|c|}{\textbf{B.2 Partial \& Early-Stop Rollout}
\textnormal{\scriptsize(Problem~B: long-tail rollout latency --- 2 of 7 shown)}}\\
\hline
Deep\-ScaleR~\cite{tan2025deepscaler} & compute ${\approx}18\times$ lower & AIME24 $28.8{\to}43.1$ & 1.5B & math & 8--32 A100 & $+$ & objective shifts across 8/16/24K stages \\ \hline
APRIL~\cite{zhou2025april} & tput $+22.5\%$ (up to $+44\%$) & quality $+2.1\%$ (up to $+8\%$) & 4B--8B & --- & --- & $+$ & reused tails are off-policy; extra requests launched \\ \hline

\rowcolor{problemB}
\multicolumn{8}{|c|}{\textbf{B.3 Speculative Decoding}
\textnormal{\scriptsize(Problem~B: long-tail rollout latency --- 2 of 13 shown)}}\\
\hline
Bubble\-Spec~\cite{xu2026bubblespec} & tput $1.4$--$1.8\times$; steps $-49$--$57\%$ & --- & --- & long-ctx reasoning & --- & \checkmark & needs idle data-parallel bubbles \\ \hline
TLT~\cite{hu2026taming} & E2E $1.7$--$2.1\times$ & --- & 7B--32B & reasoning & multi-node & \checkmark & needs idle GPUs for drafter \\ \hline

\rowcolor{problemC}
\multicolumn{8}{|c|}{\textbf{C. Rollout Selection}
\textnormal{\scriptsize(Problem~C: uninformative rollouts --- 2 of 5 shown)}}\\
\hline
PODS~\cite{xu2025not} & --- & time-to-peak $\geq 1.7\times$ & 3B--7B & math, chem & --- & $=$/$+$ & discarded rollouts already paid for \\ \hline
Rollout Replay~\cite{yoo2026rolloutreplay} & --- & $+4.35$\,pp (4B) @ same fresh rollouts & 0.6B--4B & math & --- & $+$ & age cap 10 steps; replay ratio $0.5$; fresh-anchored batches \\ \hline

\rowcolor{problemD}
\multicolumn{8}{|c|}{\textbf{D. Prompt Filtering \& Selection}
\textnormal{\scriptsize(Problem~D: uninformative prompts --- 2 of 17 shown)}}\\
\hline
GRESO~\cite{zheng2026act} & rollout $2.4\times$; train $2.0\times$ & rollouts $3.35\times$ fewer & 1.5B--32B & math & --- & $=$ & needs per-prompt reward history \\ \hline
MoPPS~\cite{qu2026prompt} & train $1.8\times$ & rollouts $-75$--$79\%$ & --- & math, countdown, geom. & --- & $\approx$/$+$ & posterior upkeep; estimates can go stale \\ \hline

\end{tabular}
\end{table}

\subsection{System Lever}

\leadin{Pipeline decoupling} methods address the structural coupling between
the rollout and training phases. In a synchronous loop the training step cannot
begin until every rollout in the batch is complete, and no new rollout can
begin until the weight update is finished, so one side of the system is idle
at any time. Pipeline decoupling lets rollout workers and training workers run
concurrently, at granularities ranging from the training step down to the
individual trajectory and out to the service interface. The defining
characteristic of the family is that the gain comes from
\emph{overlapping computation} rather than reducing the requested rollout
work; the principal cost it introduces is policy staleness, which must be
bounded and, where necessary, corrected. The family comprises
AReaL~\cite{fu2026areal}, AsyncFlow~\cite{han2025asyncflow},
StreamRL~\cite{zhong2025streamrl}, DISTFLOW~\cite{wang2025distflow},
HybridFlow~\cite{sheng2025hybridflow}, LlamaRL~\cite{wu2025llamarl},
DORA~\cite{hu2026dora}, Laminar~\cite{sheng2026laminar},
FlexMARL~\cite{jiang2026flexmarl}, RollArt~\cite{gao2026rollart},
AstraFlow~\cite{zheng2026astraflow}, StaleFlow~\cite{li2026staleflow},
and ProRL Agent~\cite{zhang2026prorl}; Section~\ref{subsec:sys-decoupling}
develops the degrees of asynchrony they span and the
staleness--throughput trade-off common to all of them.

\leadin{Resource-aware execution} methods treat the rollout--training
pipeline as a placement problem over heterogeneous or intermittently
available hardware rather than a fixed allocation over identical
accelerators. GPU generations differ in memory and throughput, spot and idle
capacity appear on changing schedules, and rollout and training have different
compute and memory profiles that no single device type serves equally well.
Methods in this family exploit that heterogeneity in three ways: harvesting
idle or spot capacity as additional rollout capacity while it is available,
placing rollout and training on device types that suit each phase best, and
replanning the split of resources between the two phases as workload demand
changes. The family is unified by changing \emph{where or from which resource
pool} a requested workload executes, rather than by changing the generation
procedure itself. Because these methods improve utilization of a fixed rollout
workload rather than reducing which work is requested, they belong to the
system lever. Its members are RLBoost~\cite{wu2025rlboost},
AReaL-Hex~\cite{yan2025arealhex}, Libra~\cite{chen2026libra},
ROSE~\cite{gao2026rose}, and BiDiRL~\cite{tan2026bidirl};
Section~\ref{subsec:sys-resource} identifies the three sources of
heterogeneity they exploit and the operating gap each one requires in order
to pay off.

\leadin{Scheduling and load balancing} methods address the allocation and
ordering of rollout requests within the available execution resources. Because
autoregressive generation is length-dependent, a batch is constrained by its
longest response: short responses finish early and leave their resources idle
while longer responses complete, creating the \emph{rollout bubble} problem.
These methods do not change the token-generation algorithm or expand the
resource pool; they change how requests are ordered, grouped, or assigned so
that execution is more balanced. Most order or group requests by predicted
response length. Some are parallelization-aware: Heddle~\cite{zhang2026heddle}
assigns greater model parallelism to predicted long-tail trajectories,
reducing their per-token latency while short trajectories use
higher-throughput, low-parallelism configurations. Others are agentic-aware:
WAR~\cite{xu2026war} places multi-turn agentic requests across rollout
replicas based on KV-cache locality and trajectory progress, while
MISA-T~\cite{hong2026misat} admits sessions from mixed workload classes under
per-class KV-cache budgets weighted by how long each class occupies the cache,
including time spent waiting for tools. The family is therefore distinguished
from resource-aware execution by its primary control variable: scheduling
changes the allocation of requested work across available workers, whereas
resource-aware execution changes the resource pool or placement regime itself.
The family reduces batching delays and memory overhead without changing the
token-generation process, policy gradient, or requested number of tokens,
making it the least invasive of the system-lever families. Other members
include RollPacker~\cite{gao2025rollpacker}, Seer~\cite{qin2026seer},
SortedRL~\cite{zhang2026sortedrl}, and TailSieve~\cite{xu2026tailsieve};
Section~\ref{subsec:sys-scheduling} formalizes this family as a makespan
minimization problem and examines the length-prediction primitive on which
many of these methods rely.

\leadin{Partial and early-stop rollout} methods reduce the number of tokens
generated per trajectory by intervening during generation rather than after it.
Not all of a rollout must be generated for it to remain useful: once a
trajectory has yielded sufficient informational content, or once the batch
contains enough complete responses, additional generation may provide limited
marginal value. Its members are APRIL~\cite{zhou2025april},
ESPO~\cite{li2026espo}, ARROL~\cite{xu2026prune},
Selective Rollout~\cite{zhai2026selective},\footnote{Selective
Rollout~\cite{zhai2026selective} is a method name. It is a different work from
GRESO~\cite{zheng2026act}, a prompt-filtering method whose paper title also
uses the phrase ``selective rollouts.''}
Kimi~k1.5~\cite{team2025kimi}, and DeepScaleR~\cite{tan2025deepscaler}.
This family sits closest to the lever boundary and is the clearest hybrid:
it is dominated by the system dimension---the prompt set and the number of
rollouts requested per prompt are unchanged, and the intervention determines
only how far an already-requested trajectory executes (rule~(iv))---but it
also reduces generated tokens and can shift the trajectory distribution, and
hence the learning signal. The taxonomy thus distinguishes the
\emph{execution decision} from the \emph{sampling decision}: truncating or
stopping a requested trajectory changes how it is executed, whereas changing
which prompts are sampled, how many rollouts are requested, or how budget is
allocated across prompts is an algorithmic intervention. The requirement that
truncated trajectories remain statistically valid for gradient estimation is
treated as a correctness constraint on the family
(Section~\ref{subsec:alg-correctness}) rather than as its defining mechanism,
which Section~\ref{subsec:sys-partial} develops.

\leadin{Speculative decoding} methods accelerate token generation itself.
Rather than producing each token through a full forward pass of the policy, a
lightweight draft mechanism proposes candidate continuations that the policy
verifies in a single parallel pass. The RL setting is distinctive because
consecutive rollout batches are not independent: the policy changes between
updates, so completions from earlier steps can provide useful draft material
for the current one. The family acts at the individual-token level, reducing
the effective cost per generated token while leaving the requested set of
trajectories unchanged. Its defining question is therefore whether the target
sampling distribution is preserved while generation is accelerated. The
family comprises RhymeRL~\cite{he2025history},
BubbleSpec~\cite{xu2026bubblespec}, DAS~\cite{shao2026beat},
SPEC-RL~\cite{liu2025specrl}, ReSpec~\cite{chen2025respec},
TLT~\cite{hu2026taming}, and SpecRoll~\cite{pham2026specroll};
Section~\ref{subsec:sys-speculative} organizes them by draft source and by
the computational cost of producing that draft.

\leadinB{System-level synthesis.}
The five system-level families optimize execution of a workload that has
already been requested, but they act on different control variables:
pipeline decoupling changes \emph{when} stages execute, resource-aware
execution changes \emph{where} they execute, scheduling changes
\emph{how requested work is assigned}, partial and early-stop rollout changes
\emph{how far an individual requested trajectory executes}, and speculative
decoding changes \emph{how tokens are generated}. These distinctions are useful
for composability: methods that remove the same execution waste may compete for
the same remaining headroom, whereas methods that act on different control
variables may retain complementary gains. Section~\ref{sec:composability}
examines these interactions explicitly.

\subsection{Algorithmic Lever}

\leadin{Rollout selection} methods operate after generation is complete and
decide which generated trajectories to use for the policy update. Rollouts are
not equally informative: those whose rewards sit near the batch mean, or for
which the model is already near-certain, produce advantage estimates close to
zero. By the time this information is available, the generation cost has
already been paid. The family therefore changes how much generated data
contributes to learning rather than how cheaply the original trajectories were
produced. It is within scope because rollout efficiency is defined by the
cost required to reach a target quality
(Equation~\eqref{eq:rollout-efficiency-problem}); post-generation selection
can reduce update computation and, when it improves learning progress, reduce
the total cost-to-quality even when the generation bill for the current batch
is unchanged. Its members are PODS~\cite{xu2025not} and
POPO~\cite{mao2026popo}; Section~\ref{subsec:alg-selection} examines the
selection criteria and their effect on the gradient estimator, and discusses
Every Rollout Counts~\cite{wang2026every}, a test-time allocation result, as
related context.

\leadin{Prompt filtering and selection} methods intervene one step earlier:
instead of choosing among \emph{completed} rollouts, they choose which
\emph{prompts} are worth generating rollouts for at all. Prompt difficulty is
highly non-uniform, and both extremes can be unproductive---prompts the policy
solves reliably and prompts it fails regardless of sampling can both yield
near-zero relative-advantage signal. The family divides into binary
filtering---applied before rollout or online during training---which skips
prompts outright, and budget allocation, which assigns different numbers of
rollouts according to estimated informativeness; difficulty prediction supplies
the signal for either decision. It is distinguished from rollout selection by
\emph{decision timing}: pre-rollout filtering can save the generation cost
itself, whereas post-rollout selection saves only update-phase computation
unless it triggers reuse or replacement. Its members are GRESO~\cite{zheng2026act},
MoPPS~\cite{qu2026prompt}, HIVE~\cite{wu2026hive},
VCRL~\cite{jiang2025vcrl}, AERO~\cite{zhang2026aero},
VIP~\cite{nguyen2026vip}, VIGOR~\cite{jiang2026vigor}, and
KGPS~\cite{zhu2026kgps}; Section~\ref{subsec:alg-prompt} organizes them by
the signal that drives the decision and by the computational cost of obtaining
that signal.

\leadinB{Algorithmic-lever synthesis.}
The two algorithmic families differ primarily in \emph{when} they decide that
rollout computation is likely to have low marginal learning value. Rollout
selection makes this decision after generation and therefore cannot recover
already-spent rollout compute unless it introduces reuse or replacement.
Prompt filtering and budget allocation make the decision before or during
generation and can therefore reduce the rollout workload itself. Both families
trade computational savings against statistical considerations because changing
which trajectories or prompts enter the update can change the effective
training distribution. These correctness and learning-efficiency trade-offs
are examined in Section~\ref{subsec:alg-correctness}.

\subsection{Bottleneck Taxonomy --- Problems, Metrics, and Resources Saved}
\label{subsec:tax-problem}

\begin{figure*}[t]
    \centering
    \captionsetup{labelfont={bf,sf,scriptsize},textfont={sf,scriptsize}}
    \makebox[\linewidth][c]{%
        \hspace*{0.01\linewidth}%
        \includegraphics[width=\textwidth]{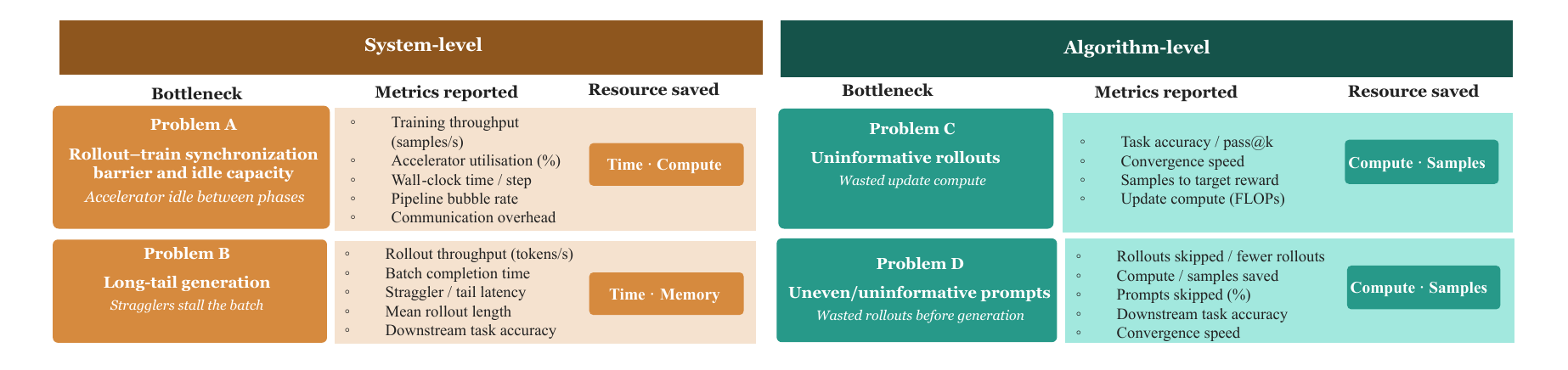}%
    }
    \caption{\textbf{Bottleneck taxonomy of rollout-efficiency methods.}}
    \label{tax:bottleneck}
\end{figure*}

The second taxonomy organizes the same body of work by the bottleneck a
method attacks. The surveyed papers target four, distinguished by where
waste arises relative to trajectory generation: (A)~rollout--train
synchronization barriers and idle capacity, between the two phases;
(B)~long-tail generation, within rollout; (C)~uninformative rollouts, after
generation; and (D)~uneven or uninformative prompts, before generation.
Figure~\ref{tax:bottleneck} presents the four as parallel columns;
Table~\ref{tab:comparison} gives the family--problem mapping in its group
headers. Problems~A and~B are \emph{execution waste}: the requested rollout
work stays in the training procedure, but time or capacity is lost
executing it. Problems~C and~D are \emph{learning-value waste}: rollout
budget is spent on trajectories or prompts with low marginal learning
value. The split complements the lever taxonomy rather than replacing it;
whether two methods actually compose depends on whether they draw on the
same resources and assumptions (Section~\ref{sec:composability}).

\leadin{Problem A --- Rollout--train synchronization barrier and idle capacity.}
In synchronous RL pipelines, training cannot start until all rollouts are
complete, and generation cannot restart until the weight update is finished, so
at any moment either the rollout workers or the training workers are idle. The
pipeline decoupling family targets this idle time directly, by removing or
relaxing the dependency that creates it. Resource-aware execution attacks the
same wasted capacity from the hardware side: RLBoost~\cite{wu2025rlboost}
harvests preemptible capacity into the rollout pool,
AReaL-Hex~\cite{yan2025arealhex} schedules rollout and training sub-tasks
across heterogeneous GPUs, Libra~\cite{chen2026libra} re-plans the
rollout/training GPU split online, ROSE~\cite{gao2026rose} borrows idle
serving GPUs, and BiDiRL~\cite{tan2026bidirl} lets either pool execute the
other's stage. Papers in this column report system-level metrics such as
training throughput, accelerator utilization, wall-clock time per training
step, pipeline bubble rate, and communication overhead. The resource recovered
is primarily time (the idle fraction of the training loop) and compute
(accelerator cycles otherwise spent waiting at a synchronization or resource
boundary).

\leadin{Problem B --- Long-tail rollout generation.}
Because RL training tasks---particularly mathematical reasoning and code
generation---have highly variable solution lengths, rollout batches mix short
responses that complete quickly with long responses that take substantially
more time. In a synchronous batch, the longest response determines completion
time while shorter responses wait with their accelerators idle, and this
straggler problem grows more severe as reasoning chains lengthen. Three
families attack different aspects of this bottleneck: scheduling and load
balancing reduces imbalance across workers, partial and early-stop rollout
reduces unnecessary continuation, and speculative decoding reduces the cost of
generating each token. They report metrics sensitive to tail behavior---rollout
throughput in tokens per second, batch completion time, straggler latency, and
mean rollout length---often alongside a downstream accuracy metric to confirm
that acceleration does not degrade policy quality. The resource saved is
primarily time (straggler wait) and, in some batching regimes, memory consumed
by padding or partially idle sequences.

\leadin{Problem C --- Uninformative rollouts (waste discovered after generation).}
Rollouts whose rewards are near the batch mean, or for which the model is
already near-certain, produce near-zero advantages and contribute little to the
update, but this is known only after the generation cost has been paid. The
rollout selection family targets this waste by reducing the number of
trajectories the policy actually uses, or by replacing uninformative groups
with informative ones already generated. The distinction from Problem~B is
therefore one of \emph{where the waste occurs}: in Problem~B the generated
trajectory is useful but expensive to execute, whereas in Problem~C the
trajectory has already been generated and is estimated to have low marginal
learning value. These papers report algorithmic quality metrics---task
accuracy, pass@k, convergence speed, and samples required to reach a target
reward---alongside or instead of system throughput, reflecting that their
primary claim is learning efficiency rather than generation speed. The
resources saved are primarily update-phase computation and, for methods that
improve learning progress, effective sample or rollout budget required to
reach a target quality.

\leadin{Problem D --- Uninformative or uneven prompts (waste avoidable before generation).}
Prompt difficulty is highly non-uniform: some prompts are solved reliably by
the current policy, others are essentially unsolvable under any reasonable
sampling budget, and both extremes can produce rollout groups with little
useful relative-advantage signal. Unlike Problem~C, where the waste is
discovered after generation, Problem~D concerns allocating rollout budget to
prompts that were unlikely to be informative under the current policy. The
prompt filtering and selection family targets this by predicting prompt
difficulty or informativeness before or during generation and adaptively
allocating, reducing, or skipping rollout budget accordingly. These papers
report a mix of efficiency and learning-quality metrics: rollout-efficiency
multipliers, compute or sample savings, the fraction of prompts skipped,
convergence speed relative to uniform sampling, and downstream accuracy to
confirm that filtering does not harm the training signal. The resource saved is
primarily compute and samples at the rollout stage itself, since generation is
avoided or reduced \emph{before} the corresponding work is fully incurred.

\leadinB{Bottleneck-lever synthesis.}
Because execution waste and learning-value waste are different resources, a
system method can improve wall-clock time without reducing rollout work,
and an algorithmic method can reduce samples without raising raw
throughput. A single efficiency number is therefore insufficient for
cross-study comparison: the relevant resource depends on where the waste
occurs. The taxonomy also exposes two properties of the empirical
literature: most methods report a gain on only one lever, leaving the other
unmeasured, and evaluation settings differ substantially in model scale,
task, workload, and hardware, so headline numbers must be read within their
measurement context rather than as a ranking. Both observations motivate
Section~\ref{sec:metrics}.

\section{System-Lever Approaches}
\label{sec:system}
\subsection{The Training Framework}
\label{subsec:sys-framework}

Every method in this section runs on top of an RL training framework, which
fixes three things: who makes control decisions, whether rollout and
training share the same devices, and how trajectories move between stages.
These choices bound what later methods can do. \emph{Control.} A single
controller for both inter-node communication and intra-node computation is
simple, but its dispatch cost grows with worker count.
HybridFlow~\cite{sheng2025hybridflow}, whose open-source system veRL is the
baseline for many rows in Table~\ref{tab:comparison}, splits the two roles;
DISTFLOW~\cite{wang2025distflow} removes the central node entirely. The two
disagree about scale, not correctness, and have not been compared at the
same large scale. \emph{Placement.} Time-sharing one GPU pool between
stages leaves nothing idle by design, but StreamRL~\cite{zhong2025streamrl}
shows the hidden cost: the stages cannot be scaled independently or given
different device types. Separate pools allow both, at the price of two
kinds of idle time---\emph{pipeline idle time} from one stage waiting on
the other, and \emph{length idle time} from long responses holding up short
ones---which Sections~\ref{subsec:sys-decoupling}
and~\ref{subsec:sys-scheduling} attack respectively; StreamRL's ablation
attributes more of its gain to removing the first than the second.
\emph{Trajectory movement.} If the trainer accepts only a full batch,
overlap happens only at batch granularity. AsyncFlow's~\cite{han2025asyncflow}
TransferQueue hands out individual samples as they finish, and its ablation
credits this path alone with much of the gain before any asynchrony is
added. AsyncFlow also hides the rollout (vLLM~\cite{kwon2023efficient},
SGLang~\cite{zheng2023sglang}) and training (FSDP~\cite{zhao2023pytorch},
Megatron-LM~\cite{shoeybi2019megatron}) engines behind service interfaces,
since those engines change faster than the RL frameworks built on them.
Table~\ref{tab:framework} (Appendix~\ref{app:framework}) places the four
frameworks on these three choices.

\subsection{Pipeline Decoupling}
\label{subsec:sys-decoupling}

\leadinB{What is the design space.}
Members of this family all relax the same barrier: training no longer waits
for a full rollout batch, and rollout no longer waits for the weight update.
They differ in how small a unit of work they let cross that barrier. At the
coarsest setting, the trainer uses a batch generated under weights at most
$k$ steps old. One-step lag ($k=1$) is the usual choice, because it is the
largest relaxation that still has a simple correction:
LlamaRL~\cite{wu2025llamarl} runs fully asynchronously at up to 405B
parameters and corrects the one step of lag with Asynchronous
Importance-weighted Policy Optimization (AIPO). A finer setting uses the
sample rather than the batch: AReaL~\cite{fu2026areal} stops rollouts when
weights update and resumes them under the new weights, so one trajectory
may span versions, and DORA~\cite{hu2026dora} extends this to several
versions at once. Laminar~\cite{sheng2026laminar} goes down to the
individual trajectory, sending weights through relay workers so no
trajectory waits on a global barrier, with lag bounded by $s \leq 4$. The
finest setting moves the barrier out of the trainer entirely:
ProRL Agent~\cite{zhang2026prorl} puts environment setup, tool calls, and
reward scoring behind an HTTP interface, so the trainer only submits tasks
and collects finished trajectories; FlexMARL~\cite{jiang2026flexmarl},
RollArt~\cite{gao2026rollart}, and AstraFlow~\cite{zheng2026astraflow} apply
the same idea to agent and multi-policy pipelines, where there is a chain of
barriers rather than a single rollout--train wait. Recent members widen the
design space on both sides of the barrier. On the system side,
RolloutPipe~\cite{chen2026rolloutpipe} shows that overlap does not require
staleness at all, pipelining disaggregated rollout and training at group
granularity while remaining strictly on-policy; Relax~\cite{zhang2026relax}
extends asynchronous training to omni-modal post-training behind a single
staleness knob; TideRL~\cite{ren2026tiderl} adds readiness-aware scheduling
with elastic worker migration for agentic workloads; and
Rollplex~\cite{lu2026rollplex} shares each GPU spatially between the two
phases while staying synchronous. On the algorithm side, a cluster of
methods raises how much lag training tolerates rather than how finely work
crosses the barrier: VCPO~\cite{huang2026vcpo} reports stability up to a lag
of $128$ steps by scaling the learning rate with the effective sample size,
$\mu$-GRPO~\cite{tian2026mugrpo} tolerates multi-stage staleness with
relaxed clipping and a negative-advantage veto, and
FlashREINFORCE~\cite{hu2026flashreinforce} and SAO~\cite{hou2026sao} remove
the group barrier entirely by training on a single rollout per prompt,
substituting a batch-mean or value-model baseline for the group baseline.

\leadinB{What does it change.}
A global synchronization point becomes a series of smaller, more frequent
handoffs, so one stage continues while the other is still running; the
price is that a single update may combine trajectories generated under
different policy versions.

\leadinB{What assumptions does it require.}
Overlap is useful only if the resulting update remains a valid learning
update. Trajectories from policies that differ substantially from the one
being optimized weaken the training signal and can destabilize learning, so
practical systems bound policy lag and, when necessary, apply off-policy
corrections; Section~\ref{subsec:alg-correctness} treats the resulting
estimator together with the other sources of off-policy error.

\leadinB{When does it help.}
The gain is bounded by the idle time that overlap can eliminate, so it is
largest where the synchronous pipeline waited most: large clusters, long and
variable trajectories, and agentic chains of barriers. There is also an
interaction with load balancing: StaleFlow~\cite{li2026staleflow} shows that
a tight lag limit restricts which trajectories may be moved between workers,
reducing the scheduler's freedom to correct length imbalance, and tracks
trajectory state globally so that both constraints can be honored at once.
Section~\ref{sec:composability} revisits this interaction.

\leadin{Synthesis.}
The primary design variable is the granularity of work allowed to cross the
rollout--train barrier, from whole batches (LlamaRL~\cite{wu2025llamarl})
through samples (AReaL~\cite{fu2026areal}, DORA~\cite{hu2026dora}) and
trajectories (Laminar~\cite{sheng2026laminar}) to the service interface
(ProRL Agent~\cite{zhang2026prorl}, FlexMARL~\cite{jiang2026flexmarl},
RollArt~\cite{gao2026rollart}, AstraFlow~\cite{zheng2026astraflow}); each
finer step buys utilization with policy lag, unless overlap is confined to
on-policy admission (RolloutPipe~\cite{chen2026rolloutpipe}). A second,
newer variable is how much lag the update itself tolerates
(VCPO~\cite{huang2026vcpo}, $\mu$-GRPO~\cite{tian2026mugrpo},
FlashREINFORCE~\cite{hu2026flashreinforce}, SAO~\cite{hou2026sao}). The
binding limitation is that the gain is capped by the idle time removed, so
once stages overlap, further asynchrony adds lag and coordination cost for
little return. The literature has not established the lag--quality curve
for any task family: systems fix one bound ($k=1$, $s\leq 4$) and report one
operating point---VCPO~\cite{huang2026vcpo}, reporting stability up to lag
$128$, is the exception---so no two methods have been compared at matched
lag.

\subsection{Resource-Aware Execution}
\label{subsec:sys-resource}

\leadinB{What is the design space.}
This family uses hardware that would otherwise be underused, matches
different stages to different device types, or reduces what each rollout
consumes on the hardware it already has. The first distinction is
\emph{where the extra capacity comes from}, because sources differ in
availability and reliability: heterogeneous device types are known in
advance and fixed, preemptible capacity is temporary and can be reclaimed,
and capacity shared with a serving workload is both temporary and contended.
A second axis is whether the rollout/training split itself is fixed or
adapted while training runs. A third changes resource demand rather than
supply: reducing the memory, precision, or auxiliary-stage cost that a
rollout occupies on whatever hardware it runs.

\leadinB{What does it change.}
Along the first two axes, the change is to \emph{where} computation runs,
not to what is computed. With \emph{different GPU types}, rollout is often
more sensitive to memory bandwidth whereas training is more sensitive to
compute and memory size, so a cluster with multiple GPU generations benefits
from assigning the two stages unevenly; AReaL-Hex~\cite{yan2025arealhex}
performs this assignment within one job. With \emph{spare capacity that can
disappear}, RLBoost~\cite{wu2025rlboost} uses the extra GPUs only for
rollout, where losing a worker affects one in-progress trajectory, rather
than for training, where it can interrupt an entire step. With \emph{spare
capacity shared with serving}, ROSE~\cite{gao2026rose} borrows idle
inference GPUs while keeping serving performance within its required
limits. Along the second axis, Libra~\cite{chen2026libra} repeatedly adjusts
the rollout/training GPU split during training and routes work through a
cost-aware queue, BiDiRL~\cite{tan2026bidirl} lets either pool execute the
other stage through a runtime that can switch roles, and
DynaResize~\cite{du2026dynaresize} replans the split online with a
hysteresis controller. The third axis shrinks what each rollout consumes.
One wing lowers numerical precision during generation: QaRL~\cite{gu2026qarl}
runs rollout with 4-bit weights, FP8-RL~\cite{qiu2026fp8rl} and
Jet-RL~\cite{xi2026jetrl} use FP8 rollout (and, in Jet-RL, FP8 training),
and AIS~\cite{zhou2026ais} keeps an FP8 sampler usable under a BF16 trainer
by adapting its importance-weight mixing per batch. A second wing reduces
the KV cache that long trajectories occupy (Sparrow~\cite{zhou2026sparrow},
Sparse-RL~\cite{luo2026sparse}, SMD~\cite{zhu2026smd}). A third moves
auxiliary stages off the critical path: AReaL-DTE~\cite{peng2026arealdte}
transmits only the small fraction of weights that change per step, and
DistRS~\cite{zhu2026distrs} serves reward evaluation from an elastic
CPU--GPU pool.

\leadinB{What assumptions does it require.}
For the capacity axes, the main requirement is operational rather than
statistical: when capacity can disappear, the system must resume, reassign,
or discard interrupted work without corrupting the training step, and when
GPUs are shared with serving, it must respect the other workload's
performance requirements. The footprint wing is the exception: quantized or
sparse rollout changes the sampling distribution itself, so these methods
carry an explicit rollout--training mismatch correction, examined with the
other mismatch cases in Section~\ref{subsec:alg-correctness}.

\leadinB{When does it help.}
The benefit depends on how much useful capacity exists beyond a fixed,
balanced allocation: heterogeneous devices help when the two stages benefit
from different hardware, and temporary or shared capacity helps when
additional machines are available part of the time. A homogeneous, fully
dedicated cluster offers much less headroom---except for footprint
reductions, which depend on hardware support (for example, FP8) rather than
on spare capacity.

\leadin{Synthesis.}
The design variable is the source of extra capacity---fixed heterogeneous
devices (AReaL-Hex~\cite{yan2025arealhex}), reclaimable preemptible GPUs
(RLBoost~\cite{wu2025rlboost}), contended serving GPUs
(ROSE~\cite{gao2026rose})---and, more recently, whether the rollout/training
split is replanned online (Libra~\cite{chen2026libra},
BiDiRL~\cite{tan2026bidirl}, DynaResize~\cite{du2026dynaresize}), or the
per-rollout footprint is reduced through lower precision, smaller KV caches,
or cheaper weight synchronization and reward serving. Gains from the
capacity axes are largest on clusters with a hardware mix or recurring spare
capacity and vanish on a homogeneous, dedicated cluster; footprint
reductions are the exception, paying off on any hardware that supports them
at the price of a sampler--trainer mismatch to correct. The principal
limitation is that every reported speedup is a property of the cluster as
much as of the method, so numbers do not transfer across environments
(Section~\ref{sec:metrics}). The literature has not measured online
repartitioning on top of an already asynchronous pipeline, the combination
Section~\ref{subsec:sys-synthesis} identifies as competing for the same idle
share.
\subsection{Scheduling and Load Balancing}
\label{subsec:sys-scheduling}

\leadinB{What is the design space.}
Given $n$ rollout requests and $m$ workers, the scheduler decides which
request goes to which worker and in what order, to minimize the time until
the entire set is finished (the \emph{makespan}). When request lengths are
known, longest-processing-time-first combined with assigning each request to
the worker that frees soonest is already close to optimal on homogeneous
machines, with approximation ratio bounded by
$4/3 - 1/(3m)$~\cite{graham1969bounds}. The challenge in rollout is that
lengths are not known in advance, so methods differ mainly in the signals
they use to predict future work and in how they respond to prediction error
and long-tail behavior.

\leadinB{What does it change.}
SortedRL~\cite{zhang2026sortedrl} and RollPacker~\cite{gao2025rollpacker}
group requests by predicted output length so that similar-length responses
share a batch, moving length variation across rather than within batches.
StreamRL~\cite{zhong2025streamrl} trains a ranker to predict output length
for dispatch; RhymeRL's HistoPipe~\cite{he2025history} uses a simpler
signal, the length distribution of the previous step, which needs no
additional model. TailSieve~\cite{xu2026tailsieve} probes the current
workload with short partial rollouts to identify groups likely to form the
long tail and keeps those groups on a separate pool that also speculates on
long outputs. Seer~\cite{qin2026seer} exploits sibling correlation within a
GRPO group: one sibling runs first as a probe whose observed length
predicts the rest, and tokens generated across siblings supply a model-free
draft for grouped speculative decoding, all while remaining synchronous and
on-policy. These results show that length imbalance can often be reduced
without abandoning on-policy execution. Scheduling becomes more complex when
the unit is a full trajectory rather than a single generation request. In
agentic RL, a trajectory may contain several generations interleaved with
tool calls, so its completion time depends on the environment as well as on
generation. Heddle~\cite{zhang2026heddle} separates trajectory time into
queueing, interference, and per-token work, addressing them with priority
scheduling, pre-sorted placement, and adaptive parallelism.
WAR~\cite{xu2026war} changes strategy with load: when the batch is small,
per-token latency is the bottleneck and suffix-tree speculation helps; when
the batch is large, batched decoding already keeps the GPU busy and
cache-aware placement becomes more useful. MISA-T~\cite{hong2026misat} moves
the problem to the shared inference service: when RLVR, RLHF, and agentic
rollouts use the same service, KV-cache capacity rather than compute can
become the limiting resource, and a scheduler based only on compute
utilization misses this contention.

\leadinB{What assumptions does it require.}
Reordering and reassigning requests do not change which trajectories are
generated, so the learning signal is unchanged as long as the same requests
remain in the same training groups. The distinction matters when scheduling
also changes batch membership: grouping by predicted length can correlate
batch composition with task difficulty, because difficulty often affects
response length. SortedRL~\cite{zhang2026sortedrl} uses this deliberately,
so that its length-sorted batches also act as a curriculum and contribute
to its reported quality gain; methods that reorder requests without
changing group membership are purely system-level optimizations.

\leadinB{When does it help.}
Scheduling has the most room to help when response lengths vary
substantially, and its benefit is erased when predicted lengths are not
more informative than the raw variation. Not every long tail is caused by
generation length, either: in agentic workloads a trajectory may be delayed
by tool calls, environment interaction, or interference from other
requests, which length-based scheduling alone cannot remove.

\leadin{Synthesis.}
The design variable is the length-prediction signal---a trained ranker
(StreamRL~\cite{zhong2025streamrl}), recent history
(RhymeRL~\cite{he2025history}), short probes
(TailSieve~\cite{xu2026tailsieve}), or sibling correlation
(Seer~\cite{qin2026seer})---together with whether the scheduler also changes
batch membership, which turns a pure system optimization into a curriculum
(SortedRL~\cite{zhang2026sortedrl}). Gains are largest under high
response-length dispersion in synchronous execution and shrink as lengths
homogenize. The principal limitation is prediction error: a predictor no more
informative than the raw length variation yields no gain, and agentic tails
driven by tool latency rather than token count fall outside length-based
scheduling altogether (Heddle~\cite{zhang2026heddle}, WAR~\cite{xu2026war},
MISA-T~\cite{hong2026misat}). The literature has not compared these
predictors on a common workload, so how gain degrades with predictor error is
unknown.
\subsection{Partial and Early-Stop Rollout}
\label{subsec:sys-partial}

\leadinB{What is the design space.}
This family limits how much of a trajectory is generated in one step.
Methods differ in whether they stop at a fixed budget, pause and resume
unfinished work, or decide dynamically that further generation is unlikely
to be useful, and in whether the decision is made per trajectory or per
group. These choices affect both the computation performed and the training
signal produced by the unfinished trajectory.

\leadinB{What does it change.}
\emph{Truncation} stops a trajectory at a fixed token budget.
DeepScaleR~\cite{tan2025deepscaler} raises this budget as training
progresses---8K context first, then 16K and 24K---because early in training
a large context limit wastes rollout compute on tokens the policy is not
yet likely to generate; it reports about $3{,}800$ A100-hours against an
estimated $70{,}000$ for one-stage training at full context.
\emph{Partial rollout reuse} also limits generation per step but does not
discard unfinished work. Kimi~k1.5~\cite{team2025kimi} stores the unfinished
continuation in a replay buffer and resumes it in a later step; only the
current segment is generated on-policy, which makes RL at 128K context
practical. APRIL~\cite{zhou2025april} applies the idea at the batch level,
starting extra requests, stopping once enough complete answers are
available, and carrying unfinished tails into later steps;
DARTS~\cite{wang2026darts} develops the same idea at larger scale, pruning
the roughly $5\%$ of trajectories that form the tail. Here the benefit
comes from avoiding the wait for the slowest request rather than from
reducing the total tokens eventually generated. \emph{Early stopping} ends a
trajectory on a learned signal rather than a fixed limit.
ESPO~\cite{li2026espo} computes a regret score from logits already produced
during sampling and stops when the smoothed score crosses a threshold
derived from the critic's value estimate; stopped trajectories are treated
as failed terminal states with an explicit terminal reward, so the negative
signal lands at the stopping point. It reduces rollout tokens by more than
$20\%$ at improved accuracy, and its ablation is an important control:
random stopping at the same rate reaches only $42.4\%$ on AIME 2024 against
$46.28\%$ for ESPO, so the benefit comes from \emph{which} trajectories are
stopped, not from stopping more of them. \emph{Group-level early stopping}
terminates generation for an entire group. Selective
Rollout~\cite{zhai2026selective} stops a group once its samples have become
similar enough that additional rollouts are unlikely to provide useful
signal, and ARROL~\cite{xu2026prune} prunes trajectories whose continued
generation is unlikely to change the learning signal and reallocates the
released capacity. POPD/TOPD~\cite{zhang2026fullrollouts} carries truncation
into on-policy \emph{distillation}, where dense teacher log-probabilities
replace the outcome reward, so prefixes alone suffice; its reported failure
on horizon-dependent tasks marks the boundary of the family, since
truncation is safe only when the learning signal does not depend on how the
trajectory ends.

\leadinB{What assumptions does it require.}
Stopping a trajectory before it finishes changes what enters training, so
each method must specify how an incomplete trajectory contributes to the
learning signal: truncation treats the shorter budget as part of the
training setup, partial reuse generates only the current segment on-policy
and may exclude older segments from the loss, early stopping turns the stop
into an explicit terminal outcome with a defined reward, and group-level
stopping raises the same question for every trajectory at once
(Section~\ref{subsec:alg-correctness}). Partial rollout reuse additionally
accumulates older trajectory segments over time, linking this family to the
policy-lag problem of Section~\ref{subsec:sys-decoupling}.

\leadinB{When does it help.}
This family works best when a trajectory's eventual usefulness can be
estimated before it finishes, as in mathematical reasoning, where an early
incorrect step makes the remainder unlikely to succeed. The opportunity is
weaker when the final reward depends strongly on a late action or
interaction, as in agentic tasks, where stopping early may discard useful
signal rather than wasted computation.

\leadin{Synthesis.}
The design variable is the stopping criterion---a fixed budget
(DeepScaleR~\cite{tan2025deepscaler}), pause-and-resume or oversample-and-prune
(Kimi~k1.5~\cite{team2025kimi}, APRIL~\cite{zhou2025april},
DARTS~\cite{wang2026darts}), a learned
per-trajectory signal (ESPO~\cite{li2026espo}), or group similarity
(Selective Rollout~\cite{zhai2026selective}, ARROL~\cite{xu2026prune})---and,
inseparably, how the incomplete trajectory enters the loss. Gains are largest
on long chain-of-thought mathematics, where an early error predicts failure
and a few stragglers set the batch time. The principal limitation is the risk
of stopping trajectories that would have succeeded, which only
ESPO~\cite{li2026espo} quantifies ($2.7\%$), together with the estimator
questions of Section~\ref{subsec:alg-correctness}. The literature has not
established transfer to agentic tasks whose reward is decided late, where
POPD/TOPD~\cite{zhang2026fullrollouts} already fails, and only ESPO reports
the random-stop control that separates the criterion's value from the value
of doing less.
\subsection{Speculative Decoding}
\label{subsec:sys-speculative}

\leadinB{What is the design space.}
Speculative decoding uses a cheaper source to propose several future tokens,
then verifies the proposals with the target policy in one pass. In RL
rollout, the main design choices are \emph{where the draft comes from} and
\emph{what it costs to produce}: a cheap drafter pays off only if its
proposals are accepted often enough, while a better drafter may reduce
target-model work but consume GPU capacity that could have served training
or rollout. \emph{Previous trajectories} are the cheapest source because RL
already produces them. SPEC-RL~\cite{liu2025specrl} reuses the trajectory
generated for the same prompt in the previous step and re-checks only the
suffix the updated policy would change. A broader approach uses a
\emph{suffix tree over past completions} (RhymeRL's
HistoSpec~\cite{he2025history}, DAS~\cite{shao2026beat},
WAR~\cite{xu2026war}); DAS reports identical training curves with and
without speculation. A \emph{trained drafter} tracks the current policy more
closely but requires additional training. ReSpec~\cite{chen2025respec}
trains a small drafter online with reward weighting;
TLT~\cite{hu2026taming} trains its drafter on otherwise idle GPUs to avoid
competing with the main job; NeMo-RL SD~\cite{iso2026accelerating} uses an
EAGLE-3 drafter initialized on domain-matched data---and finds $n$-gram
drafting slower than plain decoding in its setting---while
FastGRPO~\cite{zhang2026fastgrpo} re-tunes its draft tree to the live batch
concurrency and updates the drafter during the training phase. Another
option adds \emph{future-token heads to the policy itself}.
SpecRoll~\cite{pham2026specroll} corrects the heads' lag on two timescales,
adjusting hidden states from delayed verifier feedback and retraining head
weights only when quality keeps dropping. A related trend reuses the
\emph{multi-token prediction (MTP)} modules that DeepSeek-V3/R1 and Qwen3.5
ship natively as drafters: MTP-RL~\cite{wang2026mtprl} trains such a head
with an advantage-aware KL term that keeps acceptance from collapsing as
the policy moves, and Draft Co-Training~\cite{wang2026draftcotraining}
folds the draft loss into the trainer itself, scaling to a 122B
mixture-of-experts model. A final option needs no separate drafter at all:
EfficientRollout~\cite{kim2026efficientrollout} uses a quantized copy of the
policy to draft for its own full-precision verifier, with a roofline test
deciding when speculation pays. The cost of drafting is part of the
effective speedup. BubbleSpec~\cite{xu2026bubblespec} is the clearest
example, generating drafts in idle gaps that already exist in synchronous
data-parallel rollout, without a separate drafter or any change to
synchrony. Suffix-tree methods keep drafting cost low by running on CPU,
whereas trained drafters consume GPU time; Table~\ref{tab:draft}
(Appendix~\ref{app:draft}) summarizes the family along these two dimensions.

\leadinB{What does it change.}
Speculation changes \emph{how} tokens are generated, not the target policy
that verifies them: accepted proposals save target-model decoding steps, so
the end-to-end benefit depends on both the acceptance rate and the cost of
producing the draft.

\leadinB{What assumptions does it require.}
Proposed tokens must be likely enough under the current target policy to
make verification worthwhile. The strongest case is an \emph{exact match}: a
drafted token is kept only when it is consistent with what the target policy
would have sampled, so speculation does not change the output distribution.
BubbleSpec, SpecRoll, DAS, TLT, EfficientRollout, NeMo-RL SD, and FastGRPO
make or claim this guarantee. A weaker result is that training remains
empirically unchanged, such as similar reward curves or stable convergence
in ReSpec. SPEC-RL lies between: its reused trajectory is not generated by
the current policy, but re-checking the changed suffix reduces the
mismatch. History-based methods additionally require that useful prefixes
or prompts recur across steps, which is natural for repeated training over
a fixed prompt set but weaker when prompts are streamed or continuously
generated.

\leadinB{When does it help.}
Speculative decoding works best when the draft is cheap and the target
policy accepts a large fraction of its proposals. Acceptance falls when the
policy changes rapidly---early in training, after large updates, or under
aggressive learning schedules---and pipeline asynchrony has the same effect,
since policy lag widens the gap between the policy that produced a draft and
the policy that verifies it. Resource availability is also part of the
operating regime, which is why several members enable speculation only when
idle capacity or the current load makes it beneficial.

\leadin{Synthesis.}
The two design variables are the draft source---past trajectories
(SPEC-RL~\cite{liu2025specrl}), suffix trees (RhymeRL~\cite{he2025history},
DAS~\cite{shao2026beat}, WAR~\cite{xu2026war}), a trained drafter
(ReSpec~\cite{chen2025respec}, TLT~\cite{hu2026taming},
NeMo-RL SD~\cite{iso2026accelerating}, FastGRPO~\cite{zhang2026fastgrpo}),
heads on the policy (SpecRoll~\cite{pham2026specroll},
MTP-RL~\cite{wang2026mtprl}, Draft
Co-Training~\cite{wang2026draftcotraining}), or a quantized self-draft
(EfficientRollout~\cite{kim2026efficientrollout})---and what the drafter
costs, from free CPU lookups to continuous GPU training. Gains are largest
when the policy changes slowly, prompts recur across steps, and idle capacity
can absorb drafting (BubbleSpec~\cite{xu2026bubblespec},
TLT~\cite{hu2026taming}). The principal limitation is that acceptance falls
with policy drift, so early training, large updates, and asynchrony all erode
the speedup. The literature reports acceptance as a single average; no method
reports acceptance against training progress, so the operating range of each
drafter is unknown.
\subsection{Where These Gains Overlap}
\label{subsec:sys-synthesis}

The five families do not target five independent kinds of waste, so
stacking two mechanisms that relieve the same bottleneck yields less than
the product of their individual gains. Two such overlaps are built in.

\leadin{The tail is targeted three times.}
Scheduling, partial rollout, and speculative decoding all attack the idle
time caused by mixed response lengths. A scheduler that has balanced the
tail leaves less tail for speculation to speed up, and a stopping rule that
cuts the tail removes the tokens the scheduler was reordering. The three
families are therefore alternatives as much as complements;
TailSieve~\cite{xu2026tailsieve} and Seer~\cite{qin2026seer}, which pair
tail routing with speculation inside one system, are the only members that
measure the combination (Section~\ref{sec:composability}).

\leadin{Idle capacity is targeted twice --- but only by one wing of the
family.} Pipeline decoupling removes idle time by overlapping stages. Within
resource-aware execution, only the designs that replan the rollout/training
split (Libra~\cite{chen2026libra}, BiDiRL~\cite{tan2026bidirl},
DynaResize~\cite{du2026dynaresize}) recover that same idle share, so these
two mechanisms compete for one pool of waste: once the stages already
overlap, there is little idle capacity left to reassign, and no study has
measured the pair together. The other wings do not share this pool.
Harvesting spare or serving GPUs (RLBoost~\cite{wu2025rlboost},
ROSE~\cite{gao2026rose}) adds capacity from outside the job, matching
stages to device types (AReaL-Hex~\cite{yan2025arealhex}) exploits hardware
differences rather than idle time, and the footprint wing reduces what each
rollout consumes; all three remain useful under full asynchrony, though the
footprint wing's mismatch corrections stack importance weights on top of any
staleness correction (Section~\ref{subsec:alg-correctness}).
Section~\ref{sec:composability} quantifies the pairs for which measured
increments exist and marks the rest as hypotheses.

\section{Algorithmic-Lever Approaches}
\label{sec:algorithm}

The central source of algorithmic waste is the way group-relative RL
computes advantages. Under GRPO~\cite{shao2024deepseekmath} and related
methods, a prompt $x$ is answered by a group of $G$ sampled trajectories
whose advantages are standardized within the group
(Section~\ref{subsec:conceptual-workflow}); when every sample receives the
same reward, the group is degenerate and contributes no signal. With binary
verifiable rewards this occurs when all $G$ samples either succeed or fail:
if the policy solves $x$ with probability $p$ and samples are independent,
the probability of a degenerate group is
$P_{\text{deg}}(p,G) = p^G + (1-p)^G$, smallest at $p=1/2$ and approaching
one as $p\to 0$ or $p\to 1$. At $G=8$, a prompt solved nine times out of
ten ($p=0.9$) gives $P_{\text{deg}}\approx 0.43$; at $p=0.95$ it rises to
$\approx 0.66$, so two thirds of that prompt's groups carry no signal. Each
such group is still generated and evaluated at full cost. Two consequences
motivate the two algorithmic families. First, informativeness depends on
the \emph{prompt--policy pair}, not the prompt alone, because $p$ changes
as the policy improves. Second, whether a group will be uninformative can
be estimated only imperfectly before generation. DAPO's dynamic
sampling~\cite{yu2025dapo} is the reference approach: oversample prompts,
discard degenerate groups, and continue until the batch contains enough
with non-zero advantage. This is correct, but its rollout cost grows as
more prompts become easy. The two families below are alternatives: predict
which prompts are likely to be useful before generation, or extract more
value from generated data that would otherwise be discarded.
\subsection{Rollout Selection}
\label{subsec:alg-selection}

\leadinB{What is the design space.}
Members of this family decide after generation, when rewards are known and
the generation cost has already been paid. They differ in \emph{what is
selected}: a subset of completed trajectories, or a replacement for groups
that would otherwise be discarded.

\leadinB{What does it change.}
\emph{Selecting a subset of the generated batch.} PODS~\cite{xu2025not}
generates a large batch and trains on the maximum-reward-variance subset,
decoupling the rollout budget from the update budget: the batch is sized
for coverage while the update focuses on trajectories with more
relative-advantage signal. It reaches peak accuracy at least $1.7\times$
sooner, but the saving occurs only in the update stage, since the
discarded trajectories have already been generated. Every Rollout
Counts~\cite{wang2026every} derives the allocation of a fixed search budget
that maximizes the probability of a correct solution; we cite it as a
related allocation result, since its setting is test-time search and it
does not reduce training rollout cost. \emph{Substituting rather than
discarding.} POPO~\cite{mao2026popo} replaces degenerate groups with
prioritized off-policy groups from a replay buffer, corrected by decoupled
importance sampling, and reaches parity with oversampling baselines at
roughly $30\%$ of their rollout budget. Rollout
Replay~\cite{yoo2026rolloutreplay} and Headroom-Drift~\cite{park2026headroom}
add explicit safeguards on reuse---an age cap with fresh-anchored batches,
and a gate on measured policy drift rather than age alone.
I-PPO~\cite{shu2026right} instead derives an intrinsic stopping signal from
a held-out validation set to end training early, though the saving is
reported qualitatively.

\leadinB{What assumptions does it require.}
Post-generation selection is simplest when the selected trajectories are
already valid samples from the target policy; the concern is then the
effect of conditioning the update on the observed reward, which can reduce
estimator variance but may also remove trajectories that contain useful
signal. Replay adds a second requirement: the off-policy correction must
remain well behaved as the policy moves away from the behavior policy.

\leadinB{When does it help.}
Rollout selection is useful when generating extra samples is affordable but
using all of them in the update is wasteful; replay-based members become
attractive when the goal is to reduce new generation itself, at the cost of
off-policy correction.

\leadin{Synthesis.}
The design variable is what is selected after generation---a
maximum-variance subset (PODS~\cite{xu2025not}), a replayed replacement for
degenerate groups (POPO~\cite{mao2026popo},
Rollout Replay~\cite{yoo2026rolloutreplay},
Headroom-Drift~\cite{park2026headroom}), or an intrinsic signal for ending
training early (I-PPO~\cite{shu2026right})---which determines whether the
method saves only update compute or also new generation. Gains are largest
when oversampling is cheap relative to the update and many groups are
degenerate. The principal limitation is that post-hoc selection cannot
recover generation cost already paid, and replay recovers it only by
introducing off-policy data that must be corrected. The family has moved
from pure post-hoc selection toward guarded reuse but still lacks a shared
benchmark, so how selection gains scale with the oversampling ratio, and how
replay compares with pre-generation filtering at a matched rollout budget,
remain unmeasured.
\subsection{Prompt Filtering and Selection}
\label{subsec:alg-prompt}

\leadinB{What is the design space.}
The degeneracy condition above gives this family a common view: methods
estimate which prompts are likely to produce informative groups and direct
rollout budget toward prompts whose success probability is near $p=1/2$.
They differ in the signal used to estimate $p$, the cost of that estimate,
and whether they account for $p$ changing as the policy learns.
Table~\ref{tab:prompt-signal} (Appendix~\ref{app:prompt-signal}) summarizes
these choices.

\leadinB{What does it change.}
\emph{Binary filtering} skips prompts predicted to be uninformative. The
cheapest signal is the prompt's own reward history, refreshed as training
proceeds (GRESO~\cite{zheng2026act}, HIVE~\cite{wu2026hive},
AERO~\cite{zhang2026aero}); GRESO is representative, reporting $2.4\times$
faster rollout and up to $3.35\times$ fewer rollouts at preserved quality
($61.5$ vs.\ $61.3$). A second group fits a predictor---a Bayesian
difficulty posterior (MoPPS~\cite{qu2026prompt}), a BERT estimator
(DEPO~\cite{zhao2026depo}), a small curator model trained alongside the
actor (Actor-Curator~\cite{gu2026actorcurator}), prompt embeddings joined to
reward history (LEEPS~\cite{liang2026leeps}), or a one-time probe run
amortized across the pool (TrajVal~\cite{zhou2026beyond});
HeaPA~\cite{wang2026heapa} instead verifies pool difficulty asynchronously
with a teacher model. VCRL~\cite{jiang2025vcrl} uses group reward variance
and pairs filtering with replay. \emph{Budget allocation} gives
low-priority prompts less rather than zero budget. VIP~\cite{nguyen2026vip}
solves a convex allocation that minimizes policy-gradient variance from
Gaussian-process success estimates; VIGOR~\cite{jiang2026vigor},
HORA~\cite{wang2026hora}, and Pilot-Commit~\cite{kim2026pilotcommit}
replace the predictor with a few probe rollouts per prompt and reallocate
the remainder by observed variance---VIGOR reaches target quality with
$2.3\times$ fewer rollouts on math. KGPS~\cite{zhu2026kgps} is the clearest
treatment of the non-stationarity all these estimators face: a
linear-Gaussian state-space model whose uncertainty grows as the policy
changes, cutting rollouts by $83\%$ against dynamic sampling. Two members
stretch the family's boundary: DUET~\cite{hu2026duet} pairs allocation with
a mid-generation abort, the clearest two-family hybrid under rule~(iv) of
Section~\ref{sec:taxonomy}, and TRACE~\cite{zou2026trace} allocates below
the prompt, over turn-level prefixes in multi-turn tasks.

\leadinB{What assumptions does it require.}
Prompt filtering assumes that informativeness can be estimated well enough
before spending the full rollout budget, whether from history, a learned
predictor, a few probe samples, or a model of drift. The key difficulty is
that the target moves: $p$ is a property of the current prompt--policy
pair, not of the prompt. Permanently removing prompts also changes the
training distribution, a correctness question taken up in
Section~\ref{subsec:alg-correctness}.

\leadinB{When does it help.}
Prompt filtering and allocation are most useful when a large fraction of
rollouts produce degenerate groups---near the extremes of $p$, where
$P_{\text{deg}}$ is highest---and when informativeness can be estimated
more cheaply than generating the groups themselves. The challenge is
greatest when prompt difficulty changes quickly, because stale estimates
direct rollout budget toward the wrong prompts.

\leadin{Synthesis.}
The design variable is the informativeness signal and its cost---reward
history (GRESO~\cite{zheng2026act}), a trained predictor
(DEPO~\cite{zhao2026depo}), a few probe rollouts
(VIGOR~\cite{jiang2026vigor}), group variance with replay
(VCRL~\cite{jiang2025vcrl}), or an explicit drift model
(KGPS~\cite{zhu2026kgps})---together with whether the decision skips prompts
or reallocates budget. Gains are largest when many prompts sit at the
extremes of $p$ and their informativeness is cheaper to estimate than to
generate; dual-lever reporting is more common here than in any system-lever
family but remains a minority (5 of 17). The principal limitation is
non-stationarity: $p$ moves with the policy, so stale estimates misdirect
budget and permanent removal hides later regression. The literature has not
compared these estimators on a common workload or tested \emph{filtering}
on multi-turn tasks with partial-credit rewards, where the degenerate-group
diagnosis is less clean (Section~\ref{subsec:coverage});
TRACE~\cite{zou2026trace} allocates on multi-turn tasks but does not filter.
\subsection{Estimator Correctness Under Modified Rollout Processes}
\label{subsec:alg-correctness}

Staleness, truncation, filtering, trajectory selection, and numerical
mismatch modify the data reaching the update in different ways, but all can
be viewed through one question: \emph{does the modified rollout process
still provide a correct estimator of the gradient of the intended
objective, or does it optimize a different objective?} We consider five
cases.

\leadin{Staleness.}
Asynchronous execution generates trajectories under an older behavior policy
$\pi_{\theta_{\mathrm{old}}}$ while the update is computed for the current
policy $\pi_\theta$. The standard correction is importance weighting by the
per-token ratio $\rho$ of Section~\ref{subsec:conceptual-workflow}, so each
sampled action contributes in proportion to its likelihood under the
current policy relative to the behavior policy. Bounding staleness does not
itself make the estimator unbiased; its role is to limit the difference
between the two policies, because a larger difference makes the ratios more
variable and the corrected update noisier---which is why the systems of
Section~\ref{subsec:sys-decoupling} keep the lag small.
POPO~\cite{mao2026popo} faces the same issue from the replay side: its
decoupled importance sampling separates the distribution used for
normalization from the behavior policy that generated the replayed sample.
In both cases, correction accounts for policy mismatch but does not make
arbitrarily stale data equivalent to fresh on-policy data.

\leadin{Truncation.}
Truncating a trajectory at $L$ tokens changes the trajectory distribution
seen by the learner. The resulting update can be an unbiased estimator of
the gradient of the \emph{truncated} objective~\cite{pardo2018time}, but
generally not of the original one, so the question is whether the truncated
objective is the one the training procedure intends to optimize. The
methods of Section~\ref{subsec:sys-partial} answer differently.
DeepScaleR~\cite{tan2025deepscaler} treats the context budget as part of the
training procedure, so the objective changes as the budget increases.
ESPO~\cite{li2026espo} defines an explicit terminal failure state when a
trajectory is stopped, turning the truncated trajectory into a complete one
under the modified process; this keeps the update on-policy for that
process, but is meaningful only if the truncated state is appropriately
treated as a failure, and ESPO's reported $2.7\%$ rate of stopping
trajectories that would have succeeded measures this error directly.
Kimi~k1.5~\cite{team2025kimi} generates only the current segment on-policy
and can exclude older segments from the policy-gradient loss, avoiding the
treatment of an old continuation as fresh data at the cost of using only
part of the trajectory history.

\leadin{Filtering.}
Prompt filtering changes the distribution of prompts used for training. If
the original prompt distribution is $d(x)$ and the rule retains prompts with
probability proportional to a weight $w(x)$, the effective training
distribution becomes $\tilde{d}(x) \propto w(x)\,d(x)$, so the update targets
a different objective from uniform prompt sampling unless the change in
sampling probability is accounted for. This is particularly difficult
because the rule is policy-dependent: informativeness is estimated from
the policy-dependent success probability $p$, so the sampling rule evolves
with $\pi_\theta$ rather than defining a fixed training distribution, and
the estimated weights carry prediction error that can further distort it.
Preserved or improved quality alongside rollout savings, which most methods
report, is evidence that the reallocation is useful in the evaluated
regime, not a guarantee that the original objective is preserved. A further
concern is permanent filtering: once a prompt is removed, later regression
on it cannot be detected unless it is revisited. VCRL~\cite{jiang2025vcrl}
addresses this through replay, GRESO~\cite{zheng2026act} and
HIVE~\cite{wu2026hive} through repeated re-estimation, and
VIP~\cite{nguyen2026vip} by reducing a prompt's budget rather than removing
it. Filtering can also be read as variance reduction: concentrating budget
on the near-$p=1/2$ prompts least likely to produce degenerate groups
preserves more gradient signal per rollout, and VIP makes this explicit by
directly minimizing policy-gradient variance.

\leadin{Trajectory selection.}
Selecting completed trajectories by their realized rewards makes inclusion
in the update depend on the observed learning signal itself. Such selection
can reduce estimator variance but can also bias the update, because the
retained set is no longer a representative sample of the trajectory
distribution. The trade-off is favorable when the discarded trajectories
carry little advantage signal---the high-$P_{\text{deg}}$ setting in which
PODS~\cite{xu2025not} operates---and unfavorable when selection is
aggressive enough to remove informative ones. Whether a gain comes from the
selection criterion or merely from processing fewer trajectories is
answered by the matched-rate random control of
Section~\ref{subsec:protocols}, which ESPO's random-stop ablation
instantiates for early stopping.

\leadin{Rollout--training mismatch.}
Even a fully synchronous, on-policy pipeline can produce data that is
slightly off-policy in practice. The rollout and training engines implement
the same model with different kernels, batching, and numerical precision, so
the probability the rollout engine assigned to a sampled token is not
exactly the probability the trainer computes for it. This has the same
structure as staleness, but the gap comes from numerics rather than time, so
it does not disappear at zero lag; it is small per token but accumulates
over long trajectories~\cite{zhong2026diagnosing}. The treatments mirror
the staleness case---treat the rollout engine as the behavior policy and
correct with importance weights from its recorded probabilities, or align
the numerics of the two engines---and the quantized-rollout methods of
Section~\ref{subsec:sys-resource}, which widen the gap deliberately for
throughput, show the corrections in use: token-level importance weighting
(FP8-RL~\cite{qiu2026fp8rl}), a re-quantized learner forward pass
(QaRL~\cite{gu2026qarl}), rejection sampling with reweighting
(Sparse-RL~\cite{luo2026sparse}), and per-batch adaptation of the
importance-weight mixing from effective-sample-size diagnostics
(AIS~\cite{zhou2026ais}). Under asynchronous updates the discrepancy can be
amplified through off-policy importance weighting and may destabilize
training; ratio clipping or rejection of extreme-ratio samples is one
practical limit. Section~\ref{sec:composability} returns to this issue,
because asynchrony and reuse widen the same gap that the numerical mismatch
opens.

\leadin{Overall implication.}
These five cases differ in mechanism but expose the same trade-off: reducing
rollout cost changes either the distribution of training data or the
information available to the update. They also separate three claims that
the literature often conflates. An \emph{exact guarantee} for the intended
objective is available only where the modified process provably samples
from the target policy, as exact-match speculative decoding does
(Section~\ref{subsec:sys-speculative}). An \emph{unbiased estimator for a
modified objective} is what importance-corrected staleness, truncation
under a redefined terminal state (ESPO), and filtering under a known
reweighting achieve: the update is correct, but for an objective the method
has changed. An \emph{empirical observation that quality is preserved} is
what most filtering, selection, and mismatch results offer; it establishes
effectiveness in the evaluated regime, not estimator correctness.
Efficiency claims are meaningful only when a method states which of the
three it makes, and keeping them separate is essential when comparing
methods and when deciding whether two can safely be combined
(Section~\ref{sec:composability}).

\section[Analytical Composability]{Analytical Composability of Rollout-Efficiency Methods}
\label{sec:composability}

The taxonomy identifies the bottleneck targeted by each method, but not
whether two methods work well together. We use \emph{composability} for this
interaction: when two methods are applied together, how much of the benefit of
each remains? Table~\ref{tab:composability} presents mechanism-derived
compatibility hypotheses for every pair of families; only a small subset of
these interactions has been empirically evaluated. Of the 21 pairs, two carry
a measured incremental gain, three more are co-implemented in a single system
without an isolated measurement, and the remaining 16 are predictions from the
mechanisms and assumptions of Sections~\ref{sec:system}
and~\ref{sec:algorithm}. The table marks this status on each entry.

\leadin{Overlapping benefits ($\times$ from shared headroom).}
Some pairs are marked $\times$ because they reduce the same remaining
inefficiency, so their gains should not be expected to multiply. Decoupling and
resource-aware execution can both exploit idle capacity, though this overlap is
confined to designs that replan the rollout/training split: as
Section~\ref{subsec:sys-synthesis} shows, capacity harvesting and heterogeneous
placement draw on different headroom. Scheduling, partial and early-stop
rollout, and speculative decoding all attack the idle time caused by mixed
response lengths (Section~\ref{subsec:sys-synthesis}): a scheduler that has
balanced the tail leaves less for speculation to accelerate, and a stopping
rule that cuts the tail removes the tokens the other two were reordering or
speeding up. These pairs also carry a re-tuning obligation---a length predictor
calibrated on complete trajectories becomes inaccurate once trajectories are
truncated, so it should be calibrated under the truncation policy in use. A
similar overlap occurs between prompt filtering and rollout selection: both
reduce the contribution of uninformative groups, so if filtering has already
removed prompts whose rollouts almost always succeed or fail, fewer remain for
selection to discard.

\leadin{Conflicts ($\times$ from incompatible assumptions).}
A stronger interference occurs when one method weakens a condition the other
requires. Pipeline decoupling introduces policy staleness, whereas speculative
decoding is most effective when the drafting and verifying policies remain
close; combining the two can reduce the effectiveness of speculation even if
both provide substantial standalone speedups.
BubbleSpec's~\cite{xu2026bubblespec} use of strict synchronization can be read
as a design choice that preserves exactly the policy proximity that
speculation needs. Numerical mismatch creates another conflict: speculative
decoding combined with execution that differs across kernels or precision is
especially likely to compound it, since small differences between drafting
and verifying probabilities lower token acceptance while the rollout and
training probabilities also move apart. The footprint wing of resource-aware
execution makes this concrete: FP8-RL~\cite{qiu2026fp8rl},
QaRL~\cite{gu2026qarl}, and AIS~\cite{zhou2026ais} deliberately widen the
sampler--trainer gap for throughput and close it with importance weights, so
combining them with asynchrony stacks two corrections on one ratio. AIS's
per-batch effective-sample-size diagnostics are one way to keep the stacked
weights in range, but no method measures the combination. Staleness can
similarly constrain scheduling: under a tight bound, only some trajectories
are eligible for migration, reducing the scheduler's freedom to rebalance.
StaleFlow~\cite{li2026staleflow} co-implements the two, with a global
consistency protocol that tracks each trajectory's lifecycle and keeps
rebalancing within the staleness bound, and TideRL~\cite{ren2026tiderl} adds
readiness-aware scheduling inside a staleness-bounded pipeline; neither
isolates the increment of either mechanism. Prompt filtering can also
interfere with history-based speculation: SPEC-RL~\cite{liu2025specrl} reuses
trajectories from previous epochs for the same prompt, and an adaptive filter
changes which prompts recur across epochs, reducing these reuse
opportunities.
\begin{table}[H]
\centering
\scriptsize
\captionsetup{labelfont={bf,sf,scriptsize},textfont={sf,scriptsize}}
\renewcommand{\arraystretch}{0.92}
\setlength{\tabcolsep}{1pt}
\caption{\textbf{Pairwise composability of technique families, with evidence
status.}}
\label{tab:composability}
\linespread{0.9}\selectfont
\begin{tabular}{|
>{\raggedright\arraybackslash}m{1.9cm}|
*{6}{>{\centering\arraybackslash}m{1.3cm}|}}
\hline
\rowcolor{gray!15}
& \textbf{Decoup-} \textbf{ling}
& \textbf{Resource-} \textbf{aware}
& \textbf{Sched-} \textbf{uling}
& \textbf{Partial} \textbf{rollout}
& \textbf{Specu-} \textbf{lative}
& \textbf{Rollout} \textbf{selection} \\
\hline
Resource-aware & $(\times/\bullet^{\mathrm{m}}/\circ)^{\dag}$ & --- & $\bullet$ & $\bullet$ & $(\bullet/\circ)^{\dag}$ & $\bullet$ \\
\hline
Scheduling & $\times^{\mathrm{s}}$ & $\bullet$ & --- & $\times$ & $\times^{\mathrm{m}}$ & $\bullet$ \\
\hline
Partial rollout & $\circ^{\mathrm{s}}$ & $\bullet$ & $\times$ & --- & $\times$ & $\circ$ \\
\hline
Speculative & $\times$ & $(\bullet/\circ)^{\dag}$ & $\times^{\mathrm{m}}$ & $\times$ & --- & $\bullet$ \\
\hline
Rollout selection & $\circ$ & $\bullet$ & $\bullet$ & $\circ$ & $\bullet$ & --- \\
\hline
Prompt filtering & $\bullet$ & $\bullet$ & $\circ$ & $\bullet^{\mathrm{s}}$ & $\times$ & $\times$ \\
\hline
\end{tabular}
\vspace{2pt}
\parbox{\linewidth}{\scriptsize
\emph{Interaction:}
$\bullet$~complementary (largely independent);
$\circ$~conditional (compatible, but one constrains the other);
$\times$~overlapping or conflicting (shared headroom or incompatible
assumptions); \mbox{---}~not applicable.
\emph{Evidence:}
$^{\mathrm{m}}$~incremental gain measured on top of the other family;
$^{\mathrm{s}}$~co-implemented in a single system without an isolated
measurement; unmarked entries are mechanism-derived hypotheses.
$\dag$~The three wings of resource-aware execution
(Section~\ref{subsec:sys-resource}) interact differently: replanning the
rollout/training split shares idle headroom with decoupling ($\times$);
capacity harvesting and heterogeneous placement remain complementary under
full asynchrony ($\bullet$, measured by AReaL-Hex); and footprint reduction
(quantized, sparse, or FP8 rollout) is conditional ($\circ$) with both
decoupling and speculation, because its sampler--trainer mismatch correction
stacks on the staleness correction and a lower-precision verifier is expected
to reduce draft acceptance.
}
\end{table}
\leadinB{Conditional combinations ($\circ$).}
Some combinations remain useful, but the second method must adapt to the
conditions introduced by the first. Partial rollout reuse under pipeline
decoupling draws on one staleness budget from two sources: a resumed segment is
old because it was generated in an earlier step, and asynchrony adds lag on top
of that age, so the lag bound must account for both
(Sections~\ref{subsec:sys-partial} and~\ref{subsec:alg-correctness}).
AReaL~\cite{fu2026areal} implements exactly this combination---interrupting
rollouts at each weight update and resuming them under the new weights---but
reports no separate increment for it. Rollout reuse faces the same accounting:
a replayed group (POPO~\cite{mao2026popo},
Rollout Replay~\cite{yoo2026rolloutreplay},
Headroom-Drift~\cite{park2026headroom}) is already off-policy by its age, and
asynchrony adds lag on top, so an age cap or drift gate calibrated for
synchronous training must be tightened under decoupling. Methods that widen
the tolerated lag (VCPO~\cite{huang2026vcpo}, $\mu$-GRPO~\cite{tian2026mugrpo})
enlarge this shared budget but do not remove the need to account for both
sources. Partial rollout and rollout selection present a different issue: a
truncated trajectory carries less information about its eventual outcome than
a completed one, so if selection operates on a mixture, its criterion should
account for that difference. Prompt filtering and partial rollout are
co-implemented in DUET~\cite{hu2026duet}, which pairs cost-weighted budget
allocation with a marker-gated mid-generation abort and reports their combined
$1.62\times$ wall-clock gain rather than the increment of each. Prompt
filtering and scheduling interact through batch composition: if task
difficulty is correlated with response length, the filtered batch may become
less length-diverse, reducing the very imbalance scheduling exists to fix; the
scheduling benefit should be measured on the filtered workload. Finally,
combinations can compete for memory rather than time: replay buffers for
reuse, suffix-tree indexes for drafting, and oversampled batches for
selection each add a standing memory cost that the composed system must fit
alongside model weights and KV cache---pressure that the KV-cache reductions
of Sparrow~\cite{zhou2026sparrow} and SMD~\cite{zhu2026smd} would relieve, an
untested but mechanism-consistent pairing.

\leadinB{How to evaluate a combination.}
The common thread is that composability should be measured by the
\emph{additional} benefit of each method after the other has been applied. A
method with a large standalone speedup may add little in a combined system
because the first method already removed much of the same waste; conversely,
two methods on different bottlenecks may retain most of their individual
benefits. A simple evaluation compares four settings: the baseline, each method
alone, and both together. The combined speedup should then be compared with the
\emph{product} of the standalone speedups: a combined gain near the product
indicates complementary methods, while a gain well below it indicates
redundancy or interference. Only two entries in Table~\ref{tab:composability}
meet this standard. AReaL-Hex's~\cite{yan2025arealhex} $1.31$--$1.50\times$
gain on top of the already asynchronous AReaL~\cite{fu2026areal} is the
clearest cross-family case, and AReaL-DTE's~\cite{peng2026arealdte}
$6.8$--$7.6\times$ weight-synchronization speedup on the same asynchronous
base is a second increment in that cell, though for one stage rather than
end-to-end; TailSieve's~\cite{xu2026tailsieve} progression from $1.67\times$
(routing) to $2.59\times$ (routing plus tail speculation) is the clearest
within a single method. The remaining 19 entries are either co-implemented
without an isolated increment or untested, and measuring them---starting with
the $\times$ entries, where the hypotheses are strongest---is itself an open
problem (Section~\ref{sec:future_direction}).

\section{Evaluation, Reporting, and Comparability}
\label{sec:metrics}

The preceding sections described what the surveyed methods do and how much
they gain. This section examines the evidence itself. Its central result is
that the numbers in Table~\ref{tab:comparison} cannot always be compared
directly---not because individual measurements are wrong, but because the
literature lacks a convention for what is measured, against which baseline,
and over which portion of the training step. The first part establishes three
findings from the survey's own data
(Sections~\ref{subsec:lever-reporting}--\ref{subsec:coverage}); the second
proposes a remedy: a reporting checklist (Section~\ref{subsec:tuple}), a
metric that puts both levers on one axis (Section~\ref{subsec:cross-lever}),
and five diagnostic protocols (Section~\ref{subsec:protocols}). Throughout, a
lever counts as \emph{reported} only when a paper gives quantitative evidence
of an efficiency improvement along it; qualitative claims such as ``quality
is preserved,'' overheads, and stability ranges are evidence of
effectiveness, not of efficiency.

\subsection{Most Methods Measure Only One Efficiency Lever}
\label{subsec:lever-reporting}

Most of the surveyed literature evaluates only one of the two levers.
Table~\ref{tab:comparison} contains 80 methods: 50 report only a system
gain, 17 only an algorithmic gain, 12 both, and one (I-PPO~\cite{shu2026right})
makes an efficiency claim it does not quantify.
Table~\ref{tab:lever-reporting} breaks the counts down by family, where the
pattern is sharper still. Resource-aware execution and speculative decoding
report a throughput or wall-clock gain in every method and quantify no
reduction in rollout work for a matched learning outcome; pipeline decoupling
does the same in 15 of 17 methods, and its two exceptions
(FlashREINFORCE~\cite{hu2026flashreinforce}, SAO~\cite{hou2026sao}) invert the
pattern, reporting an algorithmic gain with no wall-clock measurement.
Rollout selection reports the reverse in four of five methods. The two
Problem~A families alone are two fifths of the survey, and 30 of their 32
methods contain no quantitative evidence of algorithmic efficiency.
\finding{1}{\textbf{Efficiency is reported one-dimensionally.} 67 of the 80
compared methods quantify only one efficiency lever---50 a system gain
without learning efficiency, 17 an algorithmic gain without wall-clock
cost---and 30 of the 32 methods in the two Problem~A families report no
quantitative algorithmic evidence at all.}

\begin{table}[h]
\centering
\scriptsize
\captionsetup{labelfont={bf,sf,scriptsize},textfont={sf,scriptsize}}
\renewcommand{\arraystretch}{0.92}
\setlength{\tabcolsep}{1pt}
\caption{\textbf{Which efficiency lever each mechanism family reports}, from
Table~\ref{tab:comparison}. \fullcircle\ every method; \halfcircle\ some
(fraction shown); \emptycircle\ none. I-PPO~\cite{shu2026right} reports an
unquantified claim and is counted on neither lever.}
\label{tab:lever-reporting}
\linespread{0.9}\selectfont
\begin{tabular}{|
>{\raggedright\arraybackslash}m{4.0cm}|
>{\centering\arraybackslash}m{1.1cm}|
>{\centering\arraybackslash}m{1.6cm}|
>{\centering\arraybackslash}m{1.8cm}|}
\hline
\rowcolor{gray!15}
\textbf{Mechanism family} & \textbf{Methods} & \textbf{System gain} &
\textbf{Algorithmic gain} \\
\hline
A.1 Pipeline decoupling & 17 & \halfcircle\ (15/17) & \halfcircle\ (2/17) \\
\hline
A.2 Resource-aware execution & 15 & \fullcircle & \emptycircle \\
\hline
B.1 Scheduling and load balancing & 7 & \fullcircle & \halfcircle\ (1/7) \\
\hline
B.2 Partial and early-stop rollout & 7 & \fullcircle & \halfcircle\ (5/7) \\
\hline
B.3 Speculative decoding & 12 & \fullcircle & \emptycircle \\
\hline
C.\phantom{1} Rollout selection & 5 & \halfcircle\ (1/5) & \halfcircle\ (4/5) \\
\hline
D.\phantom{1} Prompt filtering and selection & 17 & \halfcircle\ (5/17) &
\fullcircle \\
\hline
\end{tabular}
\end{table}

Dual reporting is concentrated where the mechanism itself crosses the lever
boundary: partial and early-stop rollout, whose stopping rules shorten
execution and change the training signal at once, and the minority of
prompt-filtering methods that report the wall-clock time recovered in
addition to the change in training distribution
(Sections~\ref{subsec:tax-approach} and~\ref{subsec:tax-problem}). Two
exceptions lie outside these families: SortedRL~\cite{zhang2026sortedrl}, the
scheduling method whose length-sorted curriculum alters the training
distribution rather than only the execution order, and
Headroom-Drift~\cite{park2026headroom}, whose drift-gated replay reports step
time alongside quality at matched fresh rollouts. The sparsity of dual-lever
evidence therefore need not mean that most techniques affect only one lever.
Papers evaluate the effect their method was designed for: pipeline-decoupling
methods report throughput even though policy lag can change the rollout work
needed to reach a given quality; 12 of the 17 prompt-filtering methods reduce
rollout work without quantifying the resulting wall-clock saving. The gap
reflects what is measured as much as what the mechanisms do.

\subsection{Five Sources of Incomparability}
\label{subsec:incomparability}

Even within the system lever, headline speedups are not comparable unless the
measurement context is aligned. Five sources of variation each change a
reported number without implying that one mechanism is more effective than
another.
(1)~\leadinB{Baseline choice.} A speedup is a ratio, and its denominator is
part of the definition. StreamRL~\cite{zhong2025streamrl} reports
$1.12$--$2.12\times$ against veRL~\cite{sheng2025hybridflow} but
$1.06$--$1.41\times$ against a colocated baseline built by the authors, on the
same system and workloads; since surveyed frameworks use different baselines,
several themselves surveyed methods, a $2\times$ gain over a naive synchronous
implementation and a $1.3\times$ gain over a tuned asynchronous one may
describe similar absolute performance.
(2)~\leadinB{Phase boundary.} Rollout-stage and end-to-end speedups are
different quantities, related by Amdahl's law only when the non-rollout phases
and their baseline are held fixed. DORA~\cite{hu2026dora} reports
$2.12\times$ end-to-end alongside up to $8.2\times$ on rollout alone;
ROSE~\cite{gao2026rose} reports $1.2$--$1.5\times$ on rollout and
$1.3$--$3.3\times$ end-to-end under different resource baselines. Both pairs
are correct and answer different questions.
(3)~\leadinB{Workload and throughput unit.} Samples per second, tokens per
second, rollout steps per second, and step time can rank two systems
differently when sequence lengths or batch composition differ: more samples
per second may mean fewer tokens per second if the samples are shorter, and a
lower step time may come from a smaller batch rather than faster execution of
equivalent work. A comparison needs both the unit and enough workload
information to fix the amount of work.
(4)~\leadinB{Aggregation statistic.} Reported gains are ranges over
configurations, and highlighting the maximum conflates best-case with typical
behavior. StaleFlow~\cite{li2026staleflow} reports $1.42$--$2.68\times$ as a
range and $1.17$--$2.01\times$ as an average; when only a maximum is given,
the reader cannot tell whether it is a typical operating point or an unusually
favorable configuration.
(5)~\leadinB{Scale and configuration.} LlamaRL~\cite{wu2025llamarl} reports
$10.7\times$ at 405B parameters, which cannot be assumed to transfer to the
7B regime most methods use, and HybridFlow~\cite{sheng2025hybridflow} reports
a $1.53$--$20.57\times$ spread across configurations of a single
system---a variation larger than the gap between many pairs of distinct
methods. A reported speedup is therefore not a standalone fact, which is why
Section~\ref{sec:composability} declines to multiply speedups when reasoning
about stacked mechanisms: without a common measurement context the product has
no interpretation.
\finding{2}{\textbf{Headline speedups are not comparable across papers.}
Baseline choice, phase boundary, throughput unit, aggregation statistic, and
scale each change a reported number without changing the mechanism; a single
system spans $1.53$--$20.57\times$ across its own configurations, a wider
range than separates most pairs of distinct methods.}
\subsection{Experimental Confounds}
\label{subsec:confounds}
Two further factors act through the experimental setting itself. (1)~\leadinB{Accelerator heterogeneity.}
The survey spans H100, H200, H800, H20, A800, A100, and GB200 GPUs and Ascend
NPUs, with AsyncFlow~\cite{han2025asyncflow} evaluated at 256 NPUs and
MISA-T~\cite{hong2026misat} on a non-NVIDIA platform. Per-device throughput is
not commensurable across these, and export-restricted variants such as H800,
A800, and H20 have reduced interconnect bandwidth in exactly the dimension
that pipeline-decoupling and resource-aware methods optimize, so a method that
removes a communication bottleneck can look more effective because of the
hardware rather than the mechanism. (2)~\leadinB{Workload structure and response length.}
Workloads differ in whether they are prefill- or decode-dominated, text-only or
multimodal, single-turn or agentic, and in reasoning length, image count, tool
calls, and environment latency; these determine where time is spent and
therefore which bottleneck an optimization can remove. A decoding optimization
that helps long-reasoning workloads may do little when prefill, multimodal
encoding, or environment interaction dominates, and the ranking of methods can
reverse when the dominant bottleneck changes. Scheduling gains in particular
grow with response-length dispersion, and speculative gains depend on
acceptance behavior (Section~\ref{sec:system}), so a paper that omits the
relevant length or acceptance statistics has not specified the conditions under
which its number holds.

\subsection{Benchmark Coverage and Its Bias}
\label{subsec:coverage}

Counting evaluation settings across Table~\ref{tab:comparison} reveals a
disclosure gap and a structural asymmetry. Eleven of 80 methods identify only
a model family or cluster configuration, not a task; ``Qwen2.5 7B/14B/32B on
up to 128 H800s'' fixes the compute environment but leaves the workload, and
hence the length distribution, unspecified. The asymmetry is in which lever
gets tested on which workload. Thirteen of the 50 system-only methods evaluate
on agentic or multi-turn workloads: FlexMARL~\cite{jiang2026flexmarl},
RollArt~\cite{gao2026rollart}, AstraFlow~\cite{zheng2026astraflow},
ProRL~Agent~\cite{zhang2026prorl}, Relax~\cite{zhang2026relax},
TideRL~\cite{ren2026tiderl}, VCPO~\cite{huang2026vcpo},
Libra~\cite{chen2026libra}, ROSE~\cite{gao2026rose},
Heddle~\cite{zhang2026heddle}, WAR~\cite{xu2026war},
MISA-T~\cite{hong2026misat}, and
Draft Co-Training~\cite{wang2026draftcotraining}. Of the 29 methods with an
algorithmic gain, six test interactive rollouts:
Selective~Rollout~\cite{zhai2026selective} and
FlashREINFORCE~\cite{hu2026flashreinforce} on ALFWorld,
HIVE~\cite{wu2026hive} on BFCL-V2 tool calling, SAO~\cite{hou2026sao} on
software-engineering tasks, Headroom-Drift~\cite{park2026headroom} on agentic
search, and TRACE~\cite{zou2026trace} on multi-hop QA and function calling.
The remaining 23, even where non-mathematical (POPO~\cite{mao2026popo} and
MoPPS~\cite{qu2026prompt} on planning and geometry, PODS~\cite{xu2025not} on
chemistry, VIGOR~\cite{jiang2026vigor} and DUET~\cite{hu2026duet} on code,
Actor-Curator~\cite{gu2026actorcurator} on puzzles), use single-turn tasks
with a verifiable final answer. This matters because algorithmic mechanisms
rest on assumptions that hold most cleanly in exactly that setting: prompt
filtering is motivated by degenerate groups, whose diagnosis is cleanest when
reward is a single terminal correct-or-incorrect signal
(Section~\ref{sec:algorithm}). In agentic tasks with partial credit and
per-step rewards, within-group reward distributions behave differently, and a
filter tuned on single-turn tasks may not transfer; TRACE, which allocates
rather than filters, is so far the only prompt-family method evaluated there.
A milder bias affects the system lever: long chain-of-thought mathematics
produces large response-length dispersion, which is precisely the condition
under which scheduling and speculative methods gain most.
\finding{3}{\textbf{The algorithmic lever is tested almost exclusively on
single-turn, verifiable tasks.} Only 6 of the 29 methods reporting an
algorithmic gain evaluate interactive rollouts, and 11 of 80 methods do not
identify a task at all, so the workload conditions under which their numbers
hold are unspecified.}
\subsection{A Minimum Reporting Checklist}
\label{subsec:tuple}

Most of the information needed to interpret an efficiency result is already
available when the experiments are run. We propose the minimum set of facts
that should accompany an efficiency claim; each addresses one of the sources
behind Findings~1--3:
(1)~\leadinB{baseline system and version}---the reference implementation,
version or commit, and whether it was tuned; when the baseline is itself a
specialized system, also report against a common reference implementation;
(2)~\leadinB{phase measured}---rollout-stage or end-to-end, and for end-to-end
claims the rollout share of baseline step time (reporting both, as
DORA~\cite{hu2026dora} and ROSE~\cite{gao2026rose} do, makes the scope
explicit);
(3)~\leadinB{throughput metric and work unit}---samples, tokens, or rollout
steps per unit time, with enough information to fix the amount and type of
work compared;
(4)~\leadinB{aggregation statistic}---whether the headline is a maximum, mean,
or median, over which configurations, and with what spread;
(5)~\leadinB{cluster and accelerator specification}---device model and count,
interconnect, and parallelism or resource-allocation strategy;
(6)~\leadinB{workload characteristics}---the task or benchmark and, for
variable-length generation, the mean, a high quantile, and the coefficient of
variation $\mathrm{CV}_L$ of response length at the stage measured, since the
ratio of length spread to mean length determines synchronization idle time;
(7)~\leadinB{lever measured}---system, algorithmic, or both; when only one is
measured, say so;
(8)~\leadinB{quality and learning-outcome evidence}---what supports any quality
or convergence claim, and whether comparisons are at matched training
progress, rollout cost, or wall-clock time.
This is a reporting convention, not a requirement for shared evaluation
infrastructure; its purpose is only to ensure an efficiency number carries
enough context to be interpreted and compared.

\subsection{A Cross-Lever Metric}
\label{subsec:cross-lever}

The checklist improves comparability within a lever but does not say how a
throughput improvement should be weighed against a reduction in required
samples. Equation~\eqref{eq:rollout-efficiency-problem} already implies the
common quantity: $C(q)$, the resource cost on the evaluation platform
required to reach target policy quality $q$, reported as a curve over $q$,
with cost in accelerator-hours when methods share a platform. Both levers
move $C(q)$---a system mechanism lowers the hours per step, an algorithmic
one lowers the steps or the work per step needed to reach $q$---so any gain
from either must appear as lower cost to the same outcome. A tokens-to-$q$
variant is a useful hardware-independent companion, though it misses
parallelization gains. Reporting a curve rather than a scalar matters
because several mechanisms trade early progress against late---curriculum-style
prompt ordering and iterative context lengthening among them---so a method
can dominate at a low threshold and lose at a high one. The metric is
measurable today: PODS~\cite{xu2025not} reports time to peak accuracy,
VIGOR~\cite{jiang2026vigor}, KGPS~\cite{zhu2026kgps}, and
Pilot-Commit~\cite{kim2026pilotcommit} report rollouts to a target,
Actor-Curator~\cite{gu2026actorcurator} and TrajVal~\cite{zhou2026beyond}
report steps to a target, HeaPA~\cite{wang2026heapa} reports PFLOPs to a
target, and DeepScaleR~\cite{tan2025deepscaler} reports compute to a stated
quality level, the clearest instance of the form.

\subsection{Diagnostic Protocols}
\label{subsec:protocols}
Five protocols would answer questions that a single headline number cannot.
Each one generalizes a pattern that already appears in a surveyed paper.(1)~\leadinB{Match the rate, vary the criterion.}
For any method that selects, filters, or truncates rollouts, compare against
an uninformed control that removes the same amount of work---random prompt
selection or random stopping---as ESPO~\cite{li2026espo} does for value-gated
early stopping. This separates the value of the criterion from the value of
simply doing less.(2)~\leadinB{Attribute combined mechanisms incrementally.}
When a method stacks mechanisms, report an ablation ladder that adds one
component at a time, as TailSieve~\cite{xu2026tailsieve} does with
$1.67\times$ for routing and $2.59\times$ for routing plus tail speculation.
The ladder shows whether components reinforce or interfere. (3)~\leadinB{Report acceptance rate against policy drift.}
Speculative acceptance may fall as the policy moves away from the draft
source, and a single average hides this. Report acceptance and the resulting
speedup at several points in training.(4)~\leadinB{Sweep staleness rather than fixing a bound.}
One staleness bound is one operating point. Report throughput, convergence,
and final quality across several bounds to show the range over which an
asynchronous method remains effective; VCPO~\cite{huang2026vcpo}, which
reports stability up to a lag of $128$, is the closest existing instance.(5)~\leadinB{Compare quality at matched cost, not only matched steps.}
Equal quality after $N$ updates establishes parity at matched steps, not
matched cost, because methods spend different time or compute per update.
Compare quality at matched wall-clock or accelerator-hours, or report the
$C(q)$ curves of Section~\ref{subsec:cross-lever}.

\section{Future Directions}
\label{sec:future_direction}
This section presents future research directions, guiding researchers to further progress this field.

\leadin{Make rollout selection an online learning problem.}
\emph{Gap.} Prompt informativeness is a property of the prompt--policy pair
(Section~\ref{sec:algorithm}), yet only KGPS~\cite{zhu2026kgps} models its
drift explicitly, the eight estimators of Table~\ref{tab:prompt-signal} have
not been compared on a common workload, and permanent filtering hides later
regression (Sections~\ref{subsec:alg-prompt} and~\ref{subsec:alg-correctness}).
\emph{Limitation.} Current methods fix an estimator and a decision rule before
training and never report how estimate quality decays as the policy moves, so
the trade between selection accuracy and the cost of fresh evidence is made
implicitly. \emph{Question.} How should a prompt's expected learning value be
estimated and updated online, and budget allocated, so that learning progress
per unit of rollout computation is maximized under non-stationarity?
\emph{Direction.} Treat selection as an online decision process---predict
value before generation, update from the current policy, allocate where
expected improvement per unit cost is highest---with an exploration term that
revisits dropped prompts and a definition of utility that reconciles reward,
gradient contribution, and training progress when they disagree.
HIVE's~\cite{wu2026hive} moving ``learning edge'' is the starting point.

\leadin{Turn generated computation into reusable computation.}
\emph{Gap.} Reuse appears in three disconnected forms---SPEC-RL's
trajectory reuse~\cite{liu2025specrl}, suffix-tree drafters, and
partial-rollout replay---each evaluated alone; Table~\ref{tab:composability}
marks speculation's interactions with decoupling and with filtering as
untested conflicts, and no method reports acceptance against policy drift
(Section~\ref{subsec:sys-speculative}). \emph{Limitation.} Each mechanism
decides reuse locally with its own staleness heuristic; none models how long
generated state stays useful as the policy changes, so reuse cannot be traded
against fresh generation. \emph{Question.} How much future rollout work can
be inferred from computation already performed, and how does that quantity
decay with policy change? \emph{Direction.} Treat trajectories, prefixes,
suffixes, KV states, and draft predictions as a computation cache, with the
system choosing among fresh, speculative, and reused generation by batch
size, model divergence, and cache state, and with reuse extending from tokens
to trajectories and multi-turn interactions without distributional bias.

\leadin{Make long-tail mitigation adaptive rather than reactive.}
\emph{Gap.} Three families attack the same tail
(Section~\ref{subsec:sys-synthesis}), yet only TailSieve~\cite{xu2026tailsieve}
reports the marginal gain of stacking two of them, length predictors have not
been compared on a common workload, and agentic tails driven by tool latency
fall outside length-based scheduling (Section~\ref{subsec:sys-scheduling}).
\emph{Limitation.} Each method commits at design time to one
response---reorder, truncate, resume (APRIL~\cite{zhou2025april}), or
speculate---chosen from predicted length alone. \emph{Question.} Given a
trajectory's progress, predicted remainder, replica load, cache locality, and
learning value, which tail action minimizes makespan and wasted computation
jointly? \emph{Direction.} An adaptive tail-management layer that decides per
trajectory among continue, migrate, isolate, restart, speculate, or defer,
evaluated with the incremental ladder of Section~\ref{subsec:protocols}
rather than by throughput or tail latency in isolation.

\leadin{Build workload-aware rollout controllers.}
\emph{Gap.} Finding~2 shows that one system spans $1.53$--$20.57\times$
across its own configurations and that resource-aware gains are bound to the
cluster; Finding~1 shows that no method's reported utility spans both levers.
\emph{Limitation.} Mechanisms are tuned independently and fixed for a run;
WAR~\cite{xu2026war}, the only method that switches mechanism by load,
switches between two. \emph{Question.} Can a controller observe workload
state and select among batching, routing, speculation, reuse, resource split,
and over-provisioning to maximize training progress per unit time?
\emph{Direction.} A common state representation (response-length
distribution, cache occupancy, concurrency, trajectory progress, load) and a
utility based on $C(q)$ (Section~\ref{subsec:cross-lever}), so the controller
decides when an algorithmic saving is worth more than system acceleration and
thereby bridges the two levers that Finding~1 shows are measured apart.

\leadin{Extend rollout efficiency from trajectories to interacting agents.}
\emph{Gap.} Finding~3: only 2 of 16 methods reporting an algorithmic gain
evaluate interactive rollouts, and the degenerate-group diagnosis that
motivates the whole algorithmic lever assumes a single terminal binary reward.
\emph{Limitation.} In agentic RL a trajectory holds multiple decisions, tool
calls, environment responses, and growing context, so neither token count nor
terminal reward measures its cost or learning value; trajectory-level signals
are inadequate for credit assignment~\cite{zeng2025multiturn}, and long
heterogeneous trajectories defeat length-based
schedulers~\cite{xu2026war}. \emph{Question.} How should rollout utility be
estimated and budget allocated under multi-step, partial-credit interaction?
\emph{Direction.} Allocate computation at the level of turns,
subtrajectories, and interactions---turn-level selection, early abandonment
and continuation, environment-aware scheduling, partial trajectory reuse, and
adaptive budgeting of the branches most likely to improve the policy---so
that the question becomes not ``which prompt should be rolled out?'' but
``which \emph{next interaction} is worth paying for?''

\leadin{Re-architect the rollout pipeline for multimodal and heterogeneous
reasoning.}
\emph{Gap.} The surveyed literature assumes text-only, decode-dominated
workloads; 11 of 43 methods do not name a task (Finding~3), and
Section~\ref{subsec:confounds} shows that method rankings reverse when
prefill, encoding, or environment interaction dominates. One systems work
addresses multimodal rollout, by overlapping visual preprocessing with
generation and sharing prefix computation~\cite{lu2026rollplex}.
\emph{Limitation.} Every cost model in Sections~\ref{sec:background}--\ref{sec:algorithm}
counts decode tokens, while vision encoding, prefill, memory footprint, and
video or interactive observations introduce cost heterogeneity that output
length does not capture. \emph{Question.} How should rollout efficiency be
defined when the unit of work may be an image, video segment, observation,
encoder state, prefill token, or decode token? \emph{Direction.} A
heterogeneous cost model supporting modality-aware batching and scheduling,
reusable multimodal representations, memory-aware routing, and selection or
compression of observations by expected learning value. These workloads also
test whether the two-lever framework is general enough: the relevant
optimization may be neither ``generate fewer tokens'' nor ``generate tokens
faster,'' but ``avoid recomputing expensive perceptual context.''

\section{Summary}
\label{sec:summary}
Rollout generation is often the dominant repeatedly incurred cost of RL for reasoning
LLMs, optimizable through two levers: a system lever that executes a given
rollout workload faster, and an algorithmic lever that reduces the rollout work
learning requires. We organized the literature by the mechanism each method
changes and the bottleneck it targets, analyzed which families compose and
which compete for the same headroom, and found that most methods quantify only
one lever while differences in baseline, phase, and workload make published
gains hard to compare. We proposed a minimum reporting convention and a
cost-to-quality metric that places both levers on a common axis, and outlined
future research directions.

\setlength{\bibsep}{0pt plus 0.3ex}
\bibliographystyle{ACM-Reference-Format}
\bibliography{ref}

\appendix
\section*{Appendix}
\section{Measurement Setup for Figure~\ref{fig:rollout-dominates}}
\label{app:measurement}
\textbf{Workloads:} The two workloads are the public multimodal inference
traces released with ModServe~\cite{qiu2025modserve}: production traffic from
an Azure multimodal serving cluster, collected over 15 to 22 October 2024 and
published as Azure LMM Inference 2025. Each record stores a timestamp, image
count, context length (text and image), and generated length. A record is
replayed as an RL prompt matching those sizes, so the run preserves a
production distribution of context and generation lengths rather than a
synthetic uniform budget. The \emph{base} workload replays the trace as
published. The \emph{long-chain-of-thought} variant keeps the same prompts
and scales generated lengths to emulate extended reasoning.
\textbf{Training configuration:} The policy is Qwen3-32B, trained with GRPO
under synchronous AReaL~\cite{fu2026areal} using a vLLM rollout
engine~\cite{kwon2023efficient}. Execution is fully synchronous: each policy
update waits for the complete rollout batch, so the phase times below are
strictly serialized and sum to the step time.
\textbf{Hardware:} Experiments run on two Ascend 910B nodes with eight NPUs
each (16 NPUs in total).
\textbf{Measured quantities:} For each of the first 60 training steps we
record wall-clock time for the four phases: rollout generation, reward
evaluation, policy update, and weight synchronization.
Figure~\ref{fig:rollout-dominates} reports the mean share of each phase over
those 60 steps for both workloads. Rollout accounts for 49\% of mean step
time on the base workload and 58\% on the long-chain-of-thought variant. The
measurements are a small-scale illustration of the trend reported at larger
scale in the
literature~\cite{gao2025rollpacker,hu2026taming,fu2026areal,zhou2025april};
they are not a benchmark of the AReaL or vLLM systems.

\section{Deployment Inference versus Rollout Generation}
\label{app:rollout-vs-inference}
Table~\ref{tab:rollout-vs-inference} expands the five properties of
Section~\ref{subsec:rollout-vs-inference}. Controlled batch arrival
covers request arrival and workload control; batch makespan covers the
primary objective and scheduling freedom; repeated and partially
predictable workloads cover request recurrence and workload
predictability; an evolving model state and training-specific
correctness correspond one-to-one to the last two rows.

\begin{table}[!htb]
\centering
\scriptsize
\captionsetup{labelfont={bf,sf,scriptsize},textfont={sf,scriptsize}}
\renewcommand{\arraystretch}{0.92}
\setlength{\tabcolsep}{1pt}
\caption{\textbf{Deployment inference versus RL rollout generation.}}
\label{tab:rollout-vs-inference}
\begin{tabular}{|
>{\raggedright\arraybackslash}m{1.8cm}|
>{\raggedright\arraybackslash}m{4.9cm}|
>{\raggedright\arraybackslash}m{5.9cm}|}
\hline
\rowcolor{gray!15}
\textbf{Property} & \textbf{Deployment inference} &
\textbf{RL rollout generation} \\
\hline
Request arrival &
Requests arrive online according to an externally determined and potentially
bursty workload. Future load is generally not known exactly. &
The trainer explicitly launches a rollout workload, typically a batch of
$B$ tasks available at the beginning of the rollout phase. \\
\hline
Workload control &
The serving system controls scheduling after requests arrive, but generally
cannot choose which requests arrive or reorder them arbitrarily across long
time scales. &
The trainer controls task sampling, batching, ordering, and often the number
of trajectories generated for each task. \\
\hline
Request recurrence &
Requests from different users are generally treated as unrelated, and the
system cannot assume that a particular request will appear again. &
Tasks are sampled repeatedly from a training distribution or finite dataset
across optimization iterations or epochs, allowing information from previous
rollouts to guide later scheduling and allocation decisions. \\
\hline
Primary objective &
Per-request latency, throughput, and service-level objectives are central;
requests should generally begin and complete promptly. &
Individual request latency is usually secondary. In synchronous training,
the primary systems objective is reducing the makespan of the complete rollout
workload that blocks the next policy update. \\
\hline
Scheduling freedom &
Deliberately delaying an admitted request typically worsens user-visible
latency or fairness. &
Tasks may be reordered, delayed, assigned different resources, truncated, or
occasionally dropped if doing so reduces overall rollout cost without harming
the learning objective. \\
\hline
Workload predictability &
Prompt lengths, output lengths, and future arrivals are only partially known
before execution. &
The batch composition is known before generation, and previous executions of
the same or similar tasks can provide estimates of trajectory length, reward,
or computational cost. \\
\hline
Model evolution &
The served model is typically fixed over many requests and changed only
occasionally through deployment updates. &
The policy changes repeatedly as training proceeds, often after every
optimization step or group of steps. Model-dependent cached state must
therefore be associated with the policy version that produced it. \\
\hline
Correctness constraint &
Numerical differences between serving implementations are primarily important
insofar as they affect returned outputs. &
Rollout outputs, log-probabilities, and other policy quantities become inputs
to optimization; inconsistencies between rollout and training engines can
directly alter the policy update. \\
\hline
\end{tabular}
\end{table}

\section{Training Frameworks}
\label{app:framework}
Table~\ref{tab:framework} places the four training frameworks discussed in
Section~\ref{subsec:sys-framework} on the three design choices---control,
placement, and trajectory movement---together with their headline gains.
These numbers appear here rather than in Table~\ref{tab:comparison} because
they use different baselines, cluster sizes, and accelerator types than the
per-method rows, and because the gain comes from the framework itself rather
than from a rollout method.

\begin{table}[!htb]
\centering
\scriptsize
\captionsetup{labelfont={bf,sf,scriptsize},textfont={sf,scriptsize}}
\renewcommand{\arraystretch}{0.92}
\setlength{\tabcolsep}{1pt}
\caption{\textbf{Training frameworks for RL post-training.}}
\label{tab:framework}
\linespread{0.9}\selectfont
\begin{tabular}{|
>{\raggedright\arraybackslash}m{2.1cm}|
>{\raggedright\arraybackslash}m{2.5cm}|
>{\raggedright\arraybackslash}m{2.3cm}|
>{\raggedright\arraybackslash}m{3.1cm}|
>{\raggedright\arraybackslash}m{2.6cm}|}
\hline
\rowcolor{gray!15}
\textbf{Framework} & \textbf{Control} & \textbf{Placement} &
\textbf{Main idea} & \textbf{Reported gain} \\
\hline
HybridFlow / veRL~\cite{sheng2025hybridflow} &
One controller between nodes, local controllers inside each node &
Shared devices, hybrid engine &
3D-HybridEngine: move the actor between training and generation layouts
without a second copy of the weights &
$1.53$--$20.57\times$ throughput (range across configurations) \\
\hline
DISTFLOW~\cite{wang2025distflow} &
Local controllers only; no central node &
Either &
Planner plus local data coordinator (cache and double buffer) &
Up to $2.63\times$; near-linear to 512 GPUs \\
\hline
StreamRL~\cite{zhong2025streamrl} &
One controller over two services &
Separate devices; can span datacenters &
Streaming generation; separates pipeline idle time from length idle time &
Up to $2.66\times$ (best configuration); $1.33\times$ cost-effectiveness \\
\hline
AsyncFlow~\cite{han2025asyncflow} &
Service interfaces; engines can be swapped &
Separate devices &
TransferQueue: hand out samples as soon as they are ready &
$1.59\times$ avg., up to $2.03\times$ (256 NPUs) \\
\hline
\end{tabular}
\end{table}

\section{Speculative Decoding for RL Rollout}
\label{app:draft}
Table~\ref{tab:draft} places the speculative decoding methods of
Table~\ref{tab:comparison} (family~B.3, plus WAR and Seer from~B.1) by draft
source, what the drafter costs to run, and the correctness guarantee each
provides.

\begin{table}[!htb]
\centering
\scriptsize
\captionsetup{labelfont={bf,sf,scriptsize},textfont={sf,scriptsize}}
\renewcommand{\arraystretch}{0.92}
\setlength{\tabcolsep}{1pt}
\caption{\textbf{Speculative decoding for RL rollout.}}
\label{tab:draft}
\linespread{0.9}\selectfont
\begin{tabular}{|
>{\raggedright\arraybackslash}m{2.4cm}|
>{\raggedright\arraybackslash}m{3.4cm}|
>{\raggedright\arraybackslash}m{2.6cm}|
>{\raggedright\arraybackslash}m{3.6cm}|}
\hline
\rowcolor{gray!15}
\textbf{Method} & \textbf{Draft source} & \textbf{Drafter cost} &
\textbf{Guarantee} \\
\hline
BubbleSpec~\cite{xu2026bubblespec} & The policy itself, run in idle
data-parallel gaps & No additional resources; uses idle GPU capacity &
Exact match; fully synchronous \\
\hline
SPEC-RL~\cite{liu2025specrl} & Previous step's trajectory for the same
prompt & Checking only ($\approx\!20$\,s/step) &
Suffix re-checked under the current policy \\
\hline
DAS~\cite{shao2026beat} & Suffix tree over past completions &
No training; runs on CPU & Exact match (verified); identical training
curves \\
\hline
RhymeRL~\cite{he2025history} & Suffix tree (HistoSpec) & No training;
runs on CPU & RL procedure unchanged \\
\hline
WAR~\cite{xu2026war} & Suffix tree, used only at low load &
No training; turned off when batching fills the GPUs & Synchronous
execution maintained \\
\hline
Seer~\cite{qin2026seer} & Tokens generated by sibling rollouts in the same
group (grouped SD) & No model; online bookkeeping only & Synchronous and
on-policy execution maintained \\
\hline
ReSpec~\cite{chen2025respec} & Small drafter, trained online with
reward weighting & Continuous GPU training & Reward convergence and
stability reported \\
\hline
TLT~\cite{hu2026taming} & Trained drafter & Trained on idle GPUs &
Exact match (lossless); accuracy preserved \\
\hline
SpecRoll~\cite{pham2026specroll} & Extra future-token heads on the policy &
Fast hidden-state fixes; head updates only when quality drops &
Exact target distribution claimed \\
\hline
Efficient\-Rollout~\cite{kim2026efficientrollout} & The policy's own weights,
quantized, as a self-drafter & Re-quantization each step
($1.3$--$2.6$\,s); no extra model; roofline test turns SD off when
compute-bound & Exact match (lossless) \\
\hline
NeMo-RL SD~\cite{iso2026accelerating} & EAGLE-3 head or native MTP heads,
synced with the policy & Draft-weight sync per step; optional online
adaptation & Exact target distribution; lossless \\
\hline
FastGRPO~\cite{zhang2026fastgrpo} & Small draft model fed by target
hidden states; draft tree resized to live concurrency & Online draft
learning inside the update phase & Rejection sampling; accuracy unchanged \\
\hline
MTP-RL~\cite{wang2026mtprl} & Parameter-sharing MTP module added to the
policy & Offline MTP warm-up, then advantage-aware KL during RL &
Verified; acceptance length grows instead of collapsing \\
\hline
Draft Co-Training~\cite{wang2026draftcotraining} & EAGLE-3 / DFlash /
DSpark drafts co-trained on stop-gradient target features & Draft loss
in the trainer; CP branch attention and PP TapChannel transport &
Rejection sampling; reward tracks the baseline up to 122B \\
\hline
\end{tabular}
\end{table}

\section{Prompt Filtering and Selection Methods}
\label{app:prompt-signal}
Table~\ref{tab:prompt-signal} places the prompt filtering and selection
methods of Table~\ref{tab:comparison} (family~D) by the signal used to
estimate prompt informativeness, what that signal costs to obtain, how it
tracks difficulty as the policy changes, and whether the resulting decision
skips prompts, adjusts their budget, or filters with replay.

\begin{table}[!htb]
\centering
\scriptsize
\captionsetup{labelfont={bf,sf,scriptsize},textfont={sf,scriptsize}}
\renewcommand{\arraystretch}{0.92}
\setlength{\tabcolsep}{1pt}
\caption{\textbf{Prompt filtering and selection methods.}}
\label{tab:prompt-signal}
\linespread{0.9}\selectfont
\begin{tabular}{|
>{\raggedright\arraybackslash}m{1.9cm}|
>{\raggedright\arraybackslash}m{3.4cm}|
>{\raggedright\arraybackslash}m{2.9cm}|
>{\raggedright\arraybackslash}m{2.6cm}|
>{\raggedright\arraybackslash}m{1.9cm}|}
\hline
\rowcolor{gray!15}
\textbf{Method} & \textbf{Signal} & \textbf{Cost of the signal} &
\textbf{Tracking policy change} & \textbf{Decision} \\
\hline
GRESO~\cite{zheng2026act} & Per-prompt reward history &
Reuses training rewards; no extra model & Repeated re-estimation &
Skip \\
\hline
MoPPS~\cite{qu2026prompt} & Bayesian estimate of prompt difficulty &
Lightweight posterior updates & Posterior updated during training &
Prioritize / skip \\
\hline
HIVE~\cite{wu2026hive} & Informativeness estimates carried across the run &
Bookkeeping only & Repeated re-estimation & Skip \\
\hline
AERO~\cite{zhang2026aero} & Statistics already collected in earlier
rollouts & No separate predictor & Statistics refresh with the run &
Skip \\
\hline
VCRL~\cite{jiang2025vcrl} & Group reward variance &
Free from current rollouts & Replay revisits filtered prompts &
Filter + replay \\
\hline
VIP~\cite{nguyen2026vip} & Gaussian-process prediction of success
probability & Predictor fit plus convex allocation solve &
Budget reduced rather than removed & Allocate \\
\hline
VIGOR~\cite{jiang2026vigor} & Observed group reward variance from initial
rollouts & A small number of probe rollouts per prompt &
Re-observed every step & Allocate \\
\hline
KGPS~\cite{zhu2026kgps} & Latent success rate in a linear-Gaussian
state-space model & Filter update per prompt &
Explicit model of drift; uncertainty grows with policy change & Allocate \\
\hline
HeaPA~\cite{wang2026heapa} & Pool difficulty verified by an asynchronous
teacher model & Teacher verification (${\approx}13\%$ of training time) plus
pool bookkeeping & Pool re-verified as training proceeds & Prioritize \\
\hline
Actor-Curator~\cite{gu2026actorcurator} & Score from a small curator model
(0.6B) trained alongside the actor & Curator training ($+14\%$ wall-clock);
two-stage candidate sampling & Curator updated online with the actor &
Prioritize / skip \\
\hline
TrajVal~\cite{zhou2026beyond} & Value estimate from a one-time probe RL run
on ${\approx}3\%$ of the pool & One probe run plus two inference passes; no
per-step cost & Static prior; not re-estimated during training &
Prioritize \\
\hline
Pilot-Commit~\cite{kim2026pilotcommit} & Reward variance from a pilot
pass that uses a fraction of the budget & Pilot rollouts; pilot lags
commit by one step & Re-estimated every step &
Allocate / skip \\
\hline
HORA~\cite{wang2026hora} & Reward variance of a probe group
($G_0{=}8$) & Probe rollouts, counted inside the budget &
Re-observed every step & Allocate \\
\hline
LEEPS~\cite{liang2026leeps} & Prompt embeddings combined with reward
history of similar prompts & Embedding lookup plus history
(${\approx}2$\,s/step) & History refreshed as rewards arrive &
Prioritize / skip \\
\hline
DEPO~\cite{zhao2026depo} & BERT estimator predicting the group
advantage before rollout & Estimator forward and online update;
100-step warm-up & Distilled from the actor's current perplexity &
Skip \\
\hline
TRACE~\cite{zou2026trace} & Predicted conditional success at prompt
roots and turn-level prefixes & Shared predictor plus tree expansion &
Predictor updated online from visited prefixes &
Allocate (roots and prefixes) \\
\hline
DUET~\cite{hu2026duet} & Running per-prompt reward-variance estimate, fed to
a cost-weighted Neyman rule & Reuses earlier steps' rollouts; one bisection
solve per batch & Running mean refreshed every step & Allocate + abort \\
\hline
\end{tabular}
\end{table}

\section{Per-Method Comparison}
\label{app:comparison}
Table~\ref{tab:comparison} gives the full per-method comparison discussed in
Section~\ref{subsec:tax-approach}. Rows are grouped by technique family, with
the bottleneck each family targets (Problems~A--D) in the group header, and
--- marks attributes absent from the original paper. The two gain columns
distinguish the levers: a \textbf{system gain} measures faster execution of a
fixed rollout workload (throughput, wall-clock time, bubble ratio, memory or
synchronization cost), whereas an \textbf{algorithmic gain} measures improved
learning efficiency (quality at a fixed rollout budget, or fewer rollouts,
steps, or compute to reach a target quality); the two capture different
effects and are not directly comparable. \textbf{Scale}, \textbf{Task}, and
\textbf{Hardware} record the original evaluation setting. \textbf{Q} records
whether policy quality was preserved, improved, comparable, or exactly
guaranteed, and \textbf{Cost / requirement} identifies what the method spends
or assumes in exchange for its gain, such as off-policy correction, rollout
over-provisioning, or dependence on heterogeneous or preemptible resources.
The 80 methods here plus the four frameworks of Table~\ref{tab:framework},
Kimi~k1.5~\cite{team2025kimi}, and Every Rollout Counts~\cite{wang2026every}
constitute the 86 included works (Section~\ref{subsec:survey-methodology}).


\begingroup
\scriptsize
\captionsetup{labelfont={bf,sf,scriptsize},textfont={sf,scriptsize}}
\renewcommand{\arraystretch}{0.78}
\setlength{\tabcolsep}{1.5pt}

\begin{longtable}{|
>{\raggedright\arraybackslash}m{1.5cm}|
>{\raggedright\arraybackslash}m{2.1cm}|
>{\raggedright\arraybackslash}m{1.9cm}|
>{\centering\arraybackslash}m{1.0cm}|
>{\raggedright\arraybackslash}m{1.4cm}|
>{\raggedright\arraybackslash}m{1.2cm}|
>{\centering\arraybackslash}m{0.5cm}|
>{\raggedright\arraybackslash}m{2.5cm}|
}
\caption{\textbf{Master comparison of surveyed rollout-efficiency methods.}
Q (quality outcome): $=$ preserved, $+$ improved, $\approx$ comparable,
\checkmark\ exact/lossless guarantee; --- not reported. Cost / requirement:
what the method spends or assumes in exchange for its gain.
Abbreviations: tput = throughput; E2E = end-to-end; util.\ = utilization;
acc.\ = accelerators; disagg.\ = disaggregated; prod.\ = production;
het.\ = heterogeneous; preempt.\ = preemptible; conv.\ = convergence;
iter.\ = iteration; MoE = mixture-of-experts; SWE = software-engineering
tasks; KV = key--value cache; OPD = on-policy distillation; ovh.\ = overhead;
SD = speculative decoding; gen.\ = generation; quant.\ = quantization;
summ.\ = summarization; pct.\ = percentile; sync = synchronous;
ESS = effective sample size; LR = learning rate; IS = importance sampling;
TIS = truncated importance sampling; W4A16/W8A8 = weight/activation bit
widths; MTP = multi-token prediction; CP/PP = context/pipeline parallelism;
TP = tensor parallelism; HBM = high-bandwidth memory; AR = autoregressive;
VLM = vision--language model.}
\label{tab:comparison}\\

\hline
\rowcolor{gray!15}
\textbf{Method} & \textbf{System gain} & \textbf{Algorithmic gain} &
\textbf{Scale} & \textbf{Task} & \textbf{Hardware} & \textbf{Q} &
\textbf{Cost / requirement} \\
\hline
\endfirsthead

\multicolumn{8}{c}{\tablename~\thetable{} -- continued from previous page}\\
\hline
\rowcolor{gray!15}
\textbf{Method} & \textbf{System gain} & \textbf{Algorithmic gain} &
\textbf{Scale} & \textbf{Task} & \textbf{Hardware} & \textbf{Q} &
\textbf{Cost / requirement} \\
\hline
\endhead

\hline
\multicolumn{8}{|r|}{\textit{Continued on next page}}\\
\hline
\endfoot

\hline
\endlastfoot

\rowcolor{problemA}
\multicolumn{8}{|c|}{\textbf{A.1 Pipeline Decoupling}
\textnormal{\scriptsize(Problem~A: idle resources from stage coupling)}}\\
\hline
AReaL~\cite{fu2026areal} & tput $2.77\times$ & --- & 1.5B--32B & math, code & --- & $=$/$+$ & stale trajectories within lag bound \\ \hline
LlamaRL~\cite{wu2025llamarl} & step $10.7\times$ (405B) & --- & 8B--405B & math & 1{,}024 H100 & $=$ & 1-step policy lag; AIPO correction \\ \hline
DORA~\cite{hu2026dora} & E2E $2.12\times$; rollout $8.2\times$ & --- & 560B MoE & reasoning & 4{,}096 acc. & --- & trajectories span multiple policy versions \\ \hline
Laminar~\cite{sheng2026laminar} & train tput $5.48\times$ & --- & --- & --- & 1{,}024 GPU & --- & lag $s\leq 4$; relay-worker weight service \\ \hline
FlexMARL~\cite{jiang2026flexmarl} & speedup $7.3\times$; util.\ $5.6\times$ & --- & 32B+14B & multi-agent & prod.\ cluster & $=$ & async micro-batch coordination \\ \hline
RollArt~\cite{gao2026rollart} & train $1.35$--$2.05\times$ & --- & 100B+ MoE & agentic & 3k+ GPU & --- & staleness bound; disaggregated hardware \\ \hline
AstraFlow~\cite{zheng2026astraflow} & train $2.7\times$ & --- & --- & math, code, agent & --- & $\approx$/$+$ & multi-policy coordination \\ \hline
StaleFlow~\cite{li2026staleflow} & tput $1.42$--$2.68\times$ & --- & --- & --- & disagg.\ cluster & $=$ & per-trajectory lifecycle tracking \\ \hline
ProRL Agent~\cite{zhang2026prorl} & tput near-linear & --- & --- & SWE, math, code & rootless HPC & $=$ & per-trajectory HTTP service hop \\ \hline
$\mu$-GRPO~\cite{tian2026mugrpo} & rollout $1.82\times$; E2E $1.53\times$ & --- & 1.7B--8B & math, code & 4 H200 & $=$/$+$ & staleness $\mu{=}1024$; relaxed clip + veto \\ \hline
Flash\-REINFORCE~\cite{hu2026flashreinforce} & --- & $+1.7$ @ half rollouts & 1.5B--30B-A3B & math, tool, ALFWorld & --- & $+$ & no group baseline; lag $4$--$8$; sequence-trust gate \\ \hline

Relax~\cite{zhang2026relax} & E2E $1.20\times$ vs.\ veRL; async $1.76$--$2.0\times$ & --- & 4B; 30B omni & omni-modal, agentic & 16 H800 & $=$ & single staleness knob; MoE routing replay at $1.9\%$ ovh. \\ \hline

Rollout\-Pipe~\cite{chen2026rolloutpipe} & rollout-to-train $-31$--$42\%$ & --- & 1.7B & reasoning, science & --- & $=$ & on-policy kept; needs disagg.\ pools; group-level admission \\ \hline

SAO~\cite{hou2026sao} & --- & $>$GRPO under async (SWE, math) & 30B-A3B & SWE, math, writing & --- & $+$ & value model doubles memory; group size 1; two-sided token clip \\ \hline

TideRL~\cite{ren2026tiderl} & goodput $5.6\times$ vs.\ sync; $1.8\times$ vs.\ AReaL & --- & 7B & agentic, multi-modal & 32 H100 & $\approx$ & staleness bound; elastic rank migration; trajectory buffer \\ \hline

VCPO~\cite{huang2026vcpo} & E2E $2.5\times$ vs.\ sync (tool-use) & stable to lag $k{=}128$ & 1.5B--7B & math, reasoning, tool-use & --- & $=$ & ESS-scaled LR; closed-form baseline adds one backward (${\approx}19\%$ step) \\ \hline

Rollplex~\cite{lu2026rollplex} & step $1.23$--$1.30\times$ vs.\ colocation; $1.57$--$2.24\times$ vs.\ disagg. & --- & 32B VLM & video reasoning & 32 H800 & $=$ & sync on-policy kept; prefix-heavy VLM only; phase-aware HBM + TP-aliased weights \\ \hline

\rowcolor{problemA}
\multicolumn{8}{|c|}{\textbf{A.2 Resource-Aware Execution}
\textnormal{\scriptsize(Problem~A: idle, mismatched, or over-provisioned hardware)}}\\
\hline
RLBoost~\cite{wu2025rlboost} & tput $1.51$--$1.97\times$ & --- & --- & --- & preempt.\ GPUs & --- & needs preemptible capacity; interruption recovery \\ \hline
AReaL-Hex~\cite{yan2025arealhex} & tput $1.31$--$1.50\times$ & --- & 1.5B--14B & --- & het.\ GPUs & $=$ & needs mixed GPU generations \\ \hline
Libra~\cite{chen2026libra} & tput $3.0\times$; conv.\ $2.5\times$ & --- & --- & agentic & 48 A800 & $=$ & elastic pool; online re-planning \\ \hline
ROSE~\cite{gao2026rose} & E2E $1.3$--$3.3\times$; rollout $1.2$--$1.5\times$ & --- & --- & agentic & shared serving & $=$ & needs idle serving GPUs; must hold SLOs \\ \hline
BiDiRL~\cite{tan2026bidirl} & tput $1.94\times$ & --- & --- & --- & 2$\times$32 GPU & $=$ & role-switching runtime \\ \hline
QaRL~\cite{gu2026qarl} & step $1.3\times$ (W4A16) & --- & 1.5B--30B-A3B & math & 8 H800 & $\approx$ & learner re-quantized to match sampler; TBPO dual clip \\ \hline
Sparse-RL~\cite{luo2026sparse} & KV $-35$--$53\%$ & --- & 1B--7B & math & 4 H20 & $\approx$ & 512-token KV budget; ${\approx}7\%$ rejected; no wall-clock reported \\ \hline

Sparrow~\cite{zhou2026sparrow} & rollout $2.0$--$2.4\times$; step $1.8$--$2.1\times$ & --- & 1.7B--14B & math, code & 4--32 H200 & $=$ & 5th-pct.\ token acceptance $\geq 0.86$; KV budget grows with model size \\ \hline

SMD~\cite{zhu2026smd} & KV $-50\%$ & --- & 1B--7B & math, summ., QA & 8 H100 & $\approx$ & mask recorded at rollout; extra dense forward for distillation \\ \hline

AReaL-DTE~\cite{peng2026arealdte} & weight sync $6.8$--$7.6\times$ ($19.9\times$ cross-cluster) & --- & 8B; 30B-A3B & math, code, logic & 8--16 H200 & \checkmark & $<2\%$ weights change/step; AdamW inversion; periodic full anchors \\ \hline

DistRS~\cite{zhu2026distrs} & reward resources $-3.79\times$ & --- & --- & CUDA code & CPU+GPU pool & $=$ & batch-level latency constraint; elastic multi-tenant service \\ \hline

Dyna\-Resize~\cite{du2026dynaresize} & tput $+66.5\%$; time $-33\%$ vs.\ best static split & --- & 8B & mixed QA, reasoning & 8 H20 & $=$ & role switch still blocks ${\approx}187$\,s (27\% hidden); hysteresis controller \\ \hline

FP8-RL~\cite{qiu2026fp8rl} & rollout tput $+44\%$ (W8A8 $+20\%$; KV FP8 $+38\%$) & --- & 8B; 30B-A3B & math & 8--32 H100 & $\approx$ & token-level TIS (clip 2); per-step re-quant.\ + weight sync; KV scale recalibration \\ \hline

Jet-RL~\cite{xi2026jetrl} & rollout $1.33\times$; train $1.41\times$; E2E $1.16\times$ & --- & 7B--32B & math & H100 & $\approx$ & FP8 for both train and rollout (${\approx}1\%$ drop); FP8-capable hardware \\ \hline

AIS~\cite{zhou2026ais} & keeps FP8 rollout $1.5$--$2.76\times$ & --- & 8B--9B & math, planning & --- & $\approx$ & per-batch ESS/KL/variance diagnostics set IS mixing weight; BF16 trainer + FP8 sampler \\ \hline

\rowcolor{problemB}
\multicolumn{8}{|c|}{\textbf{B.1 Scheduling \& Load Balancing}
\textnormal{\scriptsize(Problem~B: long-tail rollout latency)}}\\
\hline
RollPacker~\cite{gao2025rollpacker} & E2E $2.03$--$2.56\times$ & --- & 7B--32B & --- & $\leq$128 H800 & $=$ & length predictor accuracy; also $2.24\times$ vs.\ RLHFuse \\ \hline
Seer~\cite{qin2026seer} & tput $2.04\times$; tail $-72$--$94\%$ & --- & --- & production RL & --- & $=$ & probe sibling runs first per group \\ \hline
SortedRL~\cite{zhang2026sortedrl} & bubble $-50\%$+ & $+3.9$--$18.4\%$ @ fixed data & 8B--32B & --- & --- & $+$ & length-sorted batches change the training distribution \\ \hline
Heddle~\cite{zhang2026heddle} & tput $2.5\times$ & --- & 14B & agentic & --- & $=$ & runtime length prediction; trajectory migration \\ \hline
TailSieve~\cite{xu2026tailsieve} & $1.67\times$; $2.59\times$ +spec. & --- & --- & RL, distill., eval & --- & $=$ & probe rollouts; dedicated tail pool \\ \hline
WAR~\cite{xu2026war} & tput $1.4\times$/$1.6\times$ & --- & 32B & agentic & 16 H100 & $=$ & suffix-tree memory; cache-aware placement \\ \hline
MISA-T~\cite{hong2026misat} & tput $+44$--$53\%$; iter.\ $-23\%$ & --- & 35B-A3B & RLVR, RLHF, agent & non-NVIDIA & $\approx$ & admission policy only; no trainer changes \\ \hline

\rowcolor{problemB}
\multicolumn{8}{|c|}{\textbf{B.2 Partial \& Early-Stop Rollout}
\textnormal{\scriptsize(Problem~B: long-tail rollout latency)}}\\
\hline
Deep\-ScaleR~\cite{tan2025deepscaler} & compute ${\approx}18\times$ lower & AIME24 $28.8{\to}43.1$ & 1.5B & math & 8--32 A100 & $+$ & objective shifts across 8/16/24K stages \\ \hline
APRIL~\cite{zhou2025april} & tput $+22.5\%$ (up to $+44\%$) & quality $+2.1\%$ (up to $+8\%$) & 4B--8B & --- & --- & $+$ & reused tails are off-policy; extra requests launched \\ \hline
ESPO~\cite{li2026espo} & tokens $-20\%$+ & AIME24 $46.28$ vs.\ $45.25$ & 1.5B--7B & math & --- & $+$ & $2.7\%$ correct trajectories stopped; critic warmup \\ \hline
ARROL~\cite{xu2026prune} & tput $1.6$--$1.7\times$ & reward $+2.30$--$2.99$ & --- & --- & --- & $+$ & online pruning decisions per trajectory \\ \hline
Selective Rollout~\cite{zhai2026selective} & wall-clock $-10.7\%$ & success $+2.5$\,pp & 7B & ALFWorld & --- & $+$ & similarity gate computed per group \\ \hline
POPD\-/TOPD~\cite{zhang2026fullrollouts} & E2E $2.9\times$ / $-82\%$ & --- & 1.5B & math & 8 H20 & $=$/$+$ & OPD only; TOPD is $\rho{=}0.1$; truncation fails on horizon-dependent tasks \\ \hline

DARTS~\cite{wang2026darts} & E2E up to $1.77\times$ vs.\ veRL & --- & 3B--32B; 30B-A3B & math & 8--64 H20 & $=$ & over-samples $M'{>}M$; length-biased selection; ${\approx}5\%$ tails pruned \\ \hline

\rowcolor{problemB}
\multicolumn{8}{|c|}{\textbf{B.3 Speculative Decoding}
\textnormal{\scriptsize(Problem~B: long-tail rollout latency)}}\\
\hline
RhymeRL~\cite{he2025history} & E2E $2.6\times$; tput $1.86\times$ & --- & --- & math & 10s--1000s GPU & $=$ & CPU suffix-tree memory; needs recurring prompts \\ \hline
Bubble\-Spec~\cite{xu2026bubblespec} & tput $1.4$--$1.8\times$; steps $-49$--$57\%$ & --- & --- & long-ctx reasoning & --- & \checkmark & needs idle data-parallel bubbles \\ \hline
DAS~\cite{shao2026beat} & rollout $-50\%$ / $-25\%$ & --- & 1.5B--14B & math, code & $\leq$48 H100 & \checkmark & CPU drafter; needs prompt recurrence \\ \hline
SPEC-RL~\cite{liu2025specrl} & rollout $-50$--$67\%$ & --- & 1B--8B & --- & --- & $=$/$+$ & ${\approx}20$\,s/step verification; reused prefix slightly off-policy \\ \hline
ReSpec~\cite{chen2025respec} & E2E up to $4.53\times$ & --- & 3B--14B & --- & --- & $=$ & continuous GPU drafter training \\ \hline
TLT~\cite{hu2026taming} & E2E $1.7$--$2.1\times$ & --- & 7B--32B & reasoning & multi-node & \checkmark & needs idle GPUs for drafter \\ \hline
SpecRoll~\cite{pham2026specroll} & E2E $1.21$--$2.04\times$ & --- & 1.5B--14B & math & --- & \checkmark & extra heads; lag correction on two timescales \\ \hline

Efficient\-Rollout~\cite{kim2026efficientrollout} & rollout $-19.6\%$; E2E $-12.7\%$ & --- & 7B--14B & math & --- & \checkmark & self-drafter via weight quant.; roofline SD toggle; $1.3$--$2.6$\,s/step quant. \\ \hline

NeMo-RL SD~\cite{iso2026accelerating} & gen.\ $1.5$--$1.8\times$; step $1.35$--$1.41\times$ & --- & 8B & math & 32 GB200 & \checkmark & EAGLE-3 draft; domain-matched draft init; $n$-gram drafting slower than AR \\ \hline

FastGRPO~\cite{zhang2026fastgrpo} & E2E $2.35$--$2.72\times$ & --- & 1.5B--8B & math & H800 & \checkmark & draft tree re-tuned to live concurrency; online draft learning in update phase \\ \hline

MTP-RL~\cite{wang2026mtprl} & rollout $-23$--$55\%$ (max $3.15\times$) & --- & 7B--14B & math, code & --- & $\approx$ & MTP head trained from scratch; advantage-aware KL keeps acceptance from collapsing \\ \hline

Draft Co-Training~\cite{wang2026draftcotraining} & rollout $1.19$--$2.23\times$; E2E $1.16$--$1.88\times$ & --- & 8B--122B MoE & math, multi-turn & --- & $=$ & draft loss in trainer; CP branch attention + PP TapChannel; MoE verification dilutes gain \\ \hline

\rowcolor{problemC}
\multicolumn{8}{|c|}{\textbf{C. Rollout Selection}
\textnormal{\scriptsize(Problem~C: uninformative rollouts)}}\\
\hline
PODS~\cite{xu2025not} & --- & time-to-peak $\geq 1.7\times$ & 3B--7B & math, chem & --- & $=$/$+$ & discarded rollouts already paid for \\ \hline
POPO~\cite{mao2026popo} & --- & parity at ${\approx}30\%$ budget & --- & math, planning, geom. & --- & $\approx$ & replay buffer; importance-sampling correction \\ \hline
I-PPO~\cite{shu2026right} & --- & early stop (unquant.) & 1B--8B & math, physics, QA & --- & $+$ & SFT val.\ set; extra backward/episode; PPO only \\ \hline

Rollout Replay~\cite{yoo2026rolloutreplay} & --- & $+4.35$\,pp (4B) @ same fresh rollouts & 0.6B--4B & math & --- & $+$ & age cap 10 steps; replay ratio $0.5$; fresh-anchored batches \\ \hline

Headroom-Drift~\cite{park2026headroom} & step $166$ vs.\ $197$\,s (vs.\ $1.5\times$ fresh budget) & $>$GRPO @ matched fresh rollouts & --- & math, agentic search, multimodal & --- & $+$ & buffer scan per step; drift gate rejects even young groups; 7B scale check only \\ \hline

\rowcolor{problemD}
\multicolumn{8}{|c|}{\textbf{D. Prompt Filtering \& Selection}
\textnormal{\scriptsize(Problem~D: uninformative prompts)}}\\
\hline
GRESO~\cite{zheng2026act} & rollout $2.4\times$; train $2.0\times$ & rollouts $3.35\times$ fewer & 1.5B--32B & math & --- & $=$ & needs per-prompt reward history \\ \hline
MoPPS~\cite{qu2026prompt} & train $1.8\times$ & rollouts $-75$--$79\%$ & --- & math, countdown, geom. & --- & $\approx$/$+$ & posterior upkeep; estimates can go stale \\ \hline
HIVE~\cite{wu2026hive} & rollout $3.8\times$; train $2.2$--$2.3\times$ & $9.2$M fewer rollouts & 1.5B--32B & math, tool calling & --- & $=$/$+$ & cross-run bookkeeping; skipped prompts unmonitored \\ \hline
VCRL~\cite{jiang2025vcrl} & --- & $+4.67$ vs.\ GSPO (8B) & 4B--8B & math & --- & $+$ & replay buffer memory \\ \hline
AERO~\cite{zhang2026aero} & step $-45\%$ & compute $-48\%$ & 1.5B--7B & math & --- & $=$/$+$ & relies on past-rollout statistics \\ \hline
VIP~\cite{nguyen2026vip} & --- & better than uniform alloc. & --- & RLVR & --- & $+$ & GP fit + convex solve per allocation \\ \hline
VIGOR~\cite{jiang2026vigor} & --- & rollouts $2.3\times$/$1.49\times$ fewer & --- & math, code & --- & $+$ & probe rollouts per prompt \\ \hline
KGPS~\cite{zhu2026kgps} & --- & rollouts $-83\%$ & 7B & math, planning, geom. & --- & $=$/$+$ & per-prompt filter updates \\ \hline
HeaPA~\cite{wang2026heapa} & step $+2.1\%$ (ovh.) & fewer PFLOPs-to-target & 0.6B--8B & math & 96 H200 & $+$ & async teacher ${\approx}13\%$ of train time \\ \hline
Actor-Curator~\cite{gu2026actorcurator} & --- & steps $-24$--$81\%$ & 3B & math, puzzles & A100/H200 & $+$ & curator $+14\%$ wall-clock \\ \hline
TrajVal~\cite{zhou2026beyond} & --- & steps as low as $40\%$ & 1.7B--4B & math, logic & 8 A100 & $+$ & probe on ${\approx}3\%$ of pool; regime-specific prior \\ \hline

Pilot-Commit~\cite{kim2026pilotcommit} & --- & rollouts-to-target $1.5$--$1.9\times$ fewer ($2.3$--$4.0\times$ vs.\ DAPO) & 1.5B--14B & math & --- & $=$ & pilot pass lags commit by 1 step; replay buffer for overflow \\ \hline

HORA~\cite{wang2026hora} & --- & Pass@$K$ $\uparrow$ in 10/12 settings @ matched compute & 1.5B--7B & math & 3--6 A100 & $=$/$+$ & two-phase groups ($G_0{=}8$ probe); Pass@1 only comparable \\ \hline

LEEPS~\cite{liang2026leeps} & --- & $+2.6$--$3.7\%$ rel.\ vs.\ best baseline & 1.5B--7B & math & --- & $+$ & ${\approx}2$\,s/step; needs embeddings + reward history \\ \hline

DEPO~\cite{zhao2026depo} & --- & rollout cost up to $2\times$ lower; $+1.5$--$2.4\%$ acc. & 1.5B--7B & math & 16--32 H100 & $+$ & BERT difficulty estimator; 100-step warm-up before filtering \\ \hline

TRACE~\cite{zou2026trace} & --- & $+2.8$\,pts (14B, multi-hop QA) @ equal sampling cost & 8B--14B & math (tool), multi-hop QA, function calling & --- & $+$ & shared success predictor; turn-level tree expansion; multi-turn only \\ \hline

DUET~\cite{hu2026duet} & wall-clock $1.62\times$; $2.51\times$ @ half budget & $\geq$ full-budget baselines @ half budget & 1.7B--4B & math, code, QA & 8 H100 & $+$ & per-prompt variance history; marker-gated abort with $\varepsilon$-keep + importance reweighting; two thresholds ($K_1$ poll, $K_2$ 80th pct.) \\ \hline
\end{longtable}
\endgroup

\section{Survey Methodology}
\label{subsec:methodology}
We build the taxonomy in four stages---search, screen, classify,
group---summarized in Figure~\ref{fig:methodology} in the main text.

\textbf{Search.} We searched ACM Digital Library, IEEE Xplore, arXiv, and
OpenReview for work published between January~2024 and September~2026 with
the query \texttt{("rollout" OR "trajectory generation") AND
("reinforcement learning") AND ("LLM" OR "language model") AND
("efficien*" OR "throughput" OR "asynchronous")}, adapted to each
interface, and added scans of major ML and systems venues, backward and
forward snowballing from included papers, and framework release notes and
technical reports that introduce a rollout-efficiency mechanism not
otherwise represented. Database hits were pooled with the snowballed and
report-sourced records and de-duplicated to 150 candidate records.

\textbf{De-duplication.} A preprint and its published version count as one
work; we cite the published version where one exists and the most recent
preprint otherwise. Multiple papers describing one system count as one work
when they share the system name and the later paper supersedes the earlier
evaluation.

\textbf{Screen.} We kept work that meets all three inclusion criteria: it
targets \emph{training-time} trajectory generation (I1), operates in an LLM
reasoning-RL setting (I2), and changes the cost or the amount of rollout
generation (I3). We excluded serving- or test-time methods with no training
loop (E1), work on data efficiency or credit assignment with no rollout-cost
contribution (E2), and classical non-LLM RL (E3). Title and abstract
screening followed by full-text inspection against I1--I3 and E1--E3 left 86
included works. Of the 86, 80 receive comparison rows in
Table~\ref{tab:comparison}; one test-time allocation study is kept as related
context and marked as such where discussed (Section~\ref{sec:algorithm}).

\textbf{Unit of analysis.} A \emph{method} is a named technique with its own
efficiency evaluation. Framework papers (HybridFlow, DISTFLOW, StreamRL,
AsyncFlow) are included works but receive no comparison row because their
reported gains measure the framework as a whole against a different baseline
class (Section~\ref{subsec:sys-framework}, Table~\ref{tab:framework}).
Kimi~k1.5 is included but has no row because it reports no efficiency
measurement isolatable to one lever. A paper contributing two mechanisms is
assigned one primary family by rule~(iii) of Section~\ref{sec:taxonomy}.

\textbf{Classify.} We ask how each method reduces the rollout cost in
Equation~\eqref{eq:rollout-efficiency-problem}. Methods that make a fixed
rollout workload cheaper or faster take the \emph{system lever}; methods that
reduce the rollout work needed to reach the target quality $Q^*$ take the
\emph{algorithmic lever}. This separates the mechanism of improvement from
the paper's venue or application. For each method we extracted: primary
family, primary bottleneck, lever(s) with quantitative evidence, headline
gain and unit, baseline, phase measured, evaluation setting (model, task,
hardware), and quality or trade-off statement---the columns of
Table~\ref{tab:comparison}.

\textbf{Group.} Methods with similar dominant mechanisms form the seven
families used throughout the survey: pipeline decoupling, resource-aware
execution, scheduling and load balancing, partial and early-stop rollout,
speculative decoding, rollout selection, and prompt filtering and selection.
Partial and early-stop rollout sits on the boundary between the two levers,
and resource-aware execution cuts across several families. The joint
frontier is left open because it describes combinations of existing families
rather than a family of its own.


\end{document}